\documentclass[acmtog]{acmart}

\setcitestyle{square}
\setcopyright{acmcopyright}

\usepackage{booktabs} %
\usepackage[ruled]{algorithm2e} %

\SetAlFnt{\small}
\SetAlCapFnt{\small}
\SetAlCapNameFnt{\small}
\SetAlCapHSkip{0pt}

\acmJournal{TOG}
\acmYear{2019}\acmVolume{38}\acmNumber{6}\acmArticle{240}\acmMonth{11} \acmDOI{10.1145/3355089.3356573}

\usepackage{amsmath, amsthm, amsfonts, amssymb}
\usepackage{mathrsfs}
\usepackage{graphicx}
\usepackage{textcomp}
\usepackage{wrapfig}
\usepackage{subfig}
\usepackage{color}
\usepackage{xspace}
\usepackage{overpic}
\usepackage{subfig}
\usepackage{wrapfig}
\usepackage{enumitem}
\usepackage{multirow}

\usepackage{color}
\definecolor{turquoise}{cmyk}{0.65,0,0.1,0.1}
\definecolor{purple}{rgb}{0.65,0,0.65}
\definecolor{dark_green}{rgb}{0, 0.5, 0}
\definecolor{orange}{rgb}{0.8, 0.6, 0.2}
\definecolor{red}{rgb}{0.8, 0.2, 0.2}
\definecolor{brown}{rgb}{0.5, 0.16, 0.16}

\newcommand{\rh}[1]{{\color{black}#1}}
\newcommand{\ok}[1]{{\color{black}#1}}

\renewcommand{\textrightarrow}{$\rightarrow$}

\begin{document}
\title{RPM-Net: Recurrent Prediction of Motion and Parts from Point Cloud}

\author{Zihao Yan}
\authornote{Zihao Yan and Ruizhen Hu are joint first authors.}
\affiliation{%
	\institution{Shenzhen University}
}

\author{Ruizhen Hu}
\authornotemark[1]
\affiliation{%
	\institution{Shenzhen University}
}
\author{Xingguang Yan}
\affiliation{%
	\institution{Shenzhen University}
}
\author{Luanmin Chen}
\affiliation{%
	\institution{Shenzhen University}
}
\author{Oliver van Kaick}
\affiliation{%
	\institution{Carleton University}
}
\author{Hao Zhang}
\affiliation{%
	\institution{Simon Fraser University}
}
\author{Hui Huang}
\authornote{Corresponding author: Hui Huang (hhzhiyan@gmail.com)}
\affiliation{%
	\department{College of Computer Science \& Software Engineering}
	\institution{Shenzhen University}
}

\renewcommand\shortauthors{Z. Yan, R. Hu, X. Yan, L. Chen,  O. Kaick, H. Zhang, and H. Huang}

\begin{abstract}
We introduce RPM-Net, a deep learning-based approach which simultaneously infers {\em movable parts\/} and hallucinates their {\em motions\/} from a single, un-segmented, and possibly partial, 3D point cloud shape. 
RPM-Net is a novel Recurrent Neural Network (RNN), composed of an encoder-decoder pair with interleaved Long Short-Term Memory (LSTM) components, which together predict a temporal sequence of {\em pointwise displacements\/} for the input point cloud. At the same time, the displacements allow the network to learn movable parts, resulting in a motion-based shape segmentation.
Recursive applications of RPM-Net on the obtained parts can predict finer-level part motions, resulting in a hierarchical object segmentation. Furthermore, we develop a separate network to estimate part mobilities, e.g., per-part motion parameters, from the segmented motion sequence.
Both networks learn deep predictive models from a training set that exemplifies a variety of mobilities for diverse objects. 
We show results of simultaneous motion and part predictions from synthetic and real scans of 3D objects exhibiting a variety of part mobilities, possibly involving multiple movable parts.

\end{abstract}



\maketitle

\section{Introduction} \label{sec:intro}
\begin{figure}[t!]
	\vspace{-3.48cm}
	\includegraphics[width=\linewidth]{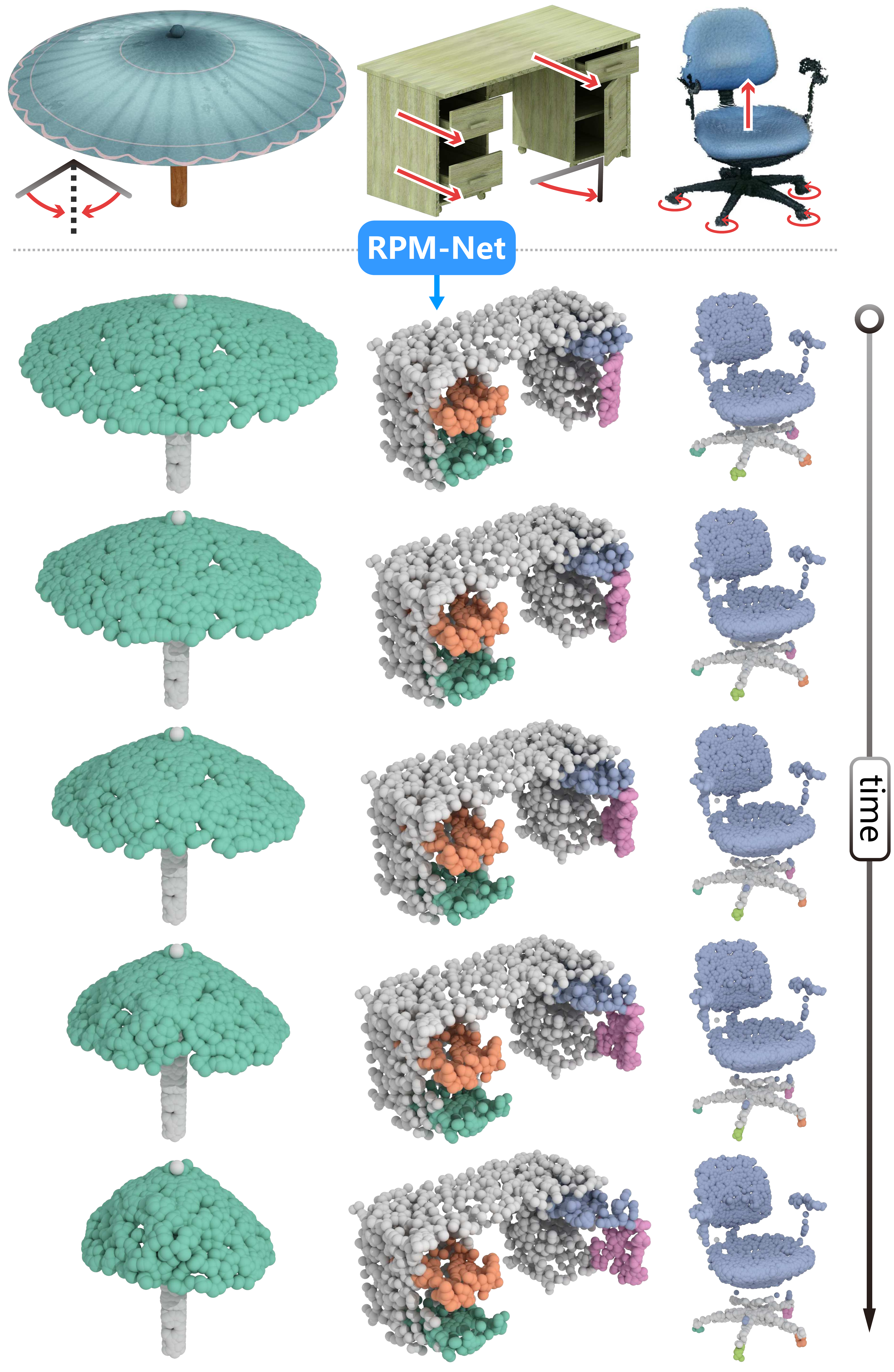}
	\caption{Given an unsegmented, possibly partial, point cloud shape, our deep recurrent neural network, RPM-Net, simultaneously hallucinates a motion sequence (via point-wise displacements) and infers a motion-based segmentation of the shape into, possibly multiple, movable parts. RPM-Net predicts a non-trivial motion for the umbrella and multi-part motions for both the cabinet (drawer sliding and door rotating) and the office chair (seat moving up and wheels rotating). The umbrella and cabinet are synthetic scans while the office chair is a {\em single-view\/} scan acquired with a Kinect sensor. Input scans to RPM-Net were downsampled to 2,048 points.}
	\label{fig:teaser}
\end{figure}

In recent years, computer graphics and related fields, such as computer vision and robotics, have devoted much attention to the inference of possible {\em motions} of 3D objects and their parts, since this problem is closely related to an understanding of {\em object affordances}~\cite{gibson79,bogoni95} and {\em functionality}~\cite{caine94,hu18}. An intriguing instance of the problem is whether and how a machine can learn to predict part motions or {\em part mobilities\/}\footnote{The subtle difference between part mobility and part motion is that the term mobility refers to the extent that a part can move; it emphasizes the geometric or transformation characteristics of part motions, e.g., the types of the motions and the reference point or axis of a rotation, etc. On the other hand, motion is a more general term which also encompasses measures reflecting physical properties, such as speed and acceleration. In our work, RPM-Net predicts movable parts and their motions simultaneously, while the second network, Mobility-Net, further predicts part mobilities.} when only few {\em static\/} states of a 3D object are given.

Hu et al.~\shortcite{hu17} introduced a data-driven approach which learns to predict part mobilities from a single static state of a 3D object, but requires the object to be well {\em segmented\/}. More recently, in deep part induction, Yi et al.~\shortcite{yi18} developed a deep neural network to infer articulated parts and their mobilities from an unsegmented object, but requires a {\em pair\/} of mobility-equivalent objects in different articulations as input. However, when functionality inference needs to be carried out over raw scans of a 3D environment, e.g., during robotic navigation, it is unrealistic to expect either object pre-segmentation or the availability of articulated object pairs.

In this paper, we introduce a deep learning-based approach which simultaneously predicts movable parts and hallucinates their motions from a {\em single\/}, {\em un-segmented\/}, and possibly {\em partial\/}, 3D point cloud shape. Hence, the key assumptions on input objects in the works of~\cite{hu17} and~\cite{yi18} are both removed. Our deep network, which is coined {\em RPM-Net\/}, is a novel Recurrent Neural Network (RNN), composed of an encoder-decoder pair with interleaved Long Short-Term Memory (LSTM) components, which together predict a temporal sequence of {\em pointwise displacements\/} for the input shape to reveal its motion. We also connect additional layers to the network to learn {\em one or more\/} movable parts from the hallucinated temporal displacement field, resulting in a motion-based shape segmentation, as shown in Figure~\ref{fig:teaser}.

Our key observation is that hallucinating and tracking pointwise movements over time represents arguably the {\em most general\/} form of motion prediction for a point cloud shape. This allows the network to process unstructured, low-level inputs and exploit the temporal characteristic of motion. At the same time, we are not making any assumption about the motion type or fitting the model to any specific motion parameters. This allows RPM-Net to learn non-trivial motions beyond simple translation and rotation, e.g., see the umbrella example in Figure~\ref{fig:teaser}. As well, the network can infer {\em multiple\/} moving parts {\em at the same time\/}, possibly with each undergoing a different type of motion; see the cabinet in Figure~\ref{fig:teaser} where the drawer slides and the door rotates about a hinge; for the office chair, the seat moves up while the wheels rotate. Note that our network can handle different sources of data, such as the synthetic scans of the umbrella and the cabinet and the {\em single-view\/} scan of the office chair via Kinect.
In addition, {\em recursive\/} applications of RPM-Net on the obtained movable parts allow prediction of finer-level part motions, resulting in a hierarchical motion-based object segmentation.

In concurrent work, Shape2Motion, Wang et al.~\shortcite{wang2019shape2motion} also aim for simultaneous motion and part prediction from an unsegmented point cloud shape. The key distinction, however, is that their network assumes that the part motion is either a translation, a rotation, or a specific combination of translation and rotation. This assumption allows the network to {\em propose\/} and then match these types of motions, as well as the motion parameters (i.e., part mobilities), based on training data. In contrast, RPM-Net makes no such assumptions and learns general shape movements. To infer part mobilities, we develop a separate network, Mobility-Net, to estimate per-part motion parameters from the output of RPM-Net. Thus, we {\em decouple\/} motion and mobility prediction.
Both RPM-Net and Mobility-Net learn their predictive models from a training set that exemplifies a variety of mobilities for diverse objects.

Our work is inspired by recent works from computer vision that predict the temporal transformation of images~\cite{zhou16,tulyakov17,xiong18}. However, besides an architecture adapted to the setting of part mobility prediction, we also introduce two important technical contributions that make the problem more tractable: (i) We introduce a loss function composed of reconstruction and motion loss components, 
which ensure that the predicted mobilities are accurate while the shape of the object is preserved. %
(ii) The use of an RNN architecture allows us to predict not only subsequent frames of a motion, but also enables us to decide when the motion has stopped. This implies that besides predicting the mobility parameters, we can also infer the {\em range\/} of a predicted mobility, e.g., how far a door can open. %

We show results of accurate and simultaneous motion and part predictions by RPM-Net from synthetic and real scans, complete or partial, of 3D objects exhibiting a variety of part movements, possibly involving multiple movable parts. We validate the components of our approach and compare our method to baseline approaches. 
In addition, we compare results of part mobility prediction by RPM-Net + Mobility-Net, to Shape2Motion, demonstrating both the generality and higher level of accuracy of our method.
Finally, we show results of hierarchical motion prediction.

\section{Related work} \label{sec:related}

Methods have been proposed to acquire and reconstruct objects along with their motion~\cite{tevs12,stuckler15,li16}, represent and understand object motion~\cite{hermans2013,pirk2017}, and even predict part mobilities from static objects~\cite{hu17,yi18}. The motivation behind these efforts is that a more complete understanding of object motion can be used for graphics applications such as animation, object pose modification, and reconstruction, as well as robotics applications such as the modeling of agent-object interactions in 3D environments.
In this section, we discuss previous works most related to the task of mobility inference for objects and their parts.

\subsection{Affordance analysis} 
In robotics, considerable work has focused on the problem of affordance detection, where the goal is to identify regions of an object that afford certain interactions, e.g., grasping or pushing~\cite{hassanin18}. Recent approaches employ deep networks for labeling images with affordance labels~\cite{roy16}, or physics-based simulations to derive human utilities closely related to affordances~\cite{zhu16}. Although affordance detection identifies regions that can undergo certain types of motion such as rolling or sliding, the detected motions are described only with a label and are limited to interactions of an agent. Thus, they do not represent general motions that an object can undergo. 

More general approaches for affordance analysis are based on the idea of {\em human pose hallucination}, where a human pose that best fits the context of a given scene is predicted to aid in understanding the scene~\cite{jiang13}. Human pose hallucination can also be used to infer the functional category of an object, based on how a human agent interacts with the object~\cite{kim14}. Closely related to affordance and human pose analysis is {\em activity recognition}, one example being the detection of activity regions in an input scene, which are regions that support specific types of human activities such as having a meal or watching TV~\cite{savva14}. The focus of these approaches is on understanding at a high-level the actions that can be carried out with certain objects or in given scenes, while the specific motions or part mobilities related to these actions are not detected nor modeled by these methods.

\subsection{Temporal transformation of images}
In computer vision, methods have been proposed to infer the state of an object in a future time, based on a depiction of the object in the present. These methods implicitly predict the motion that the objects in an image undergo and extrapolate the motion to a future time. A common solution is to generate the future frames of an input image with generative adversarial networks (GANs) trained on video data~ \cite{zhou16,xiong18}. On the other hand, Tulyakov et al.~\shortcite{tulyakov17} explicitly learn to decompose a video into content and motion components, which can then be used to create future frames of a video according to selected content and motion. Moreover, Chao et al.~\shortcite{chao17} introduce a 3D pose forecasting network to infer the future state of the pose of human models detected on images.

Similar to these approaches, we also introduce a learning-based approach based on deep networks for motion inference. However, we formulate the problem as segmenting input geometry and predicting the motion of movable parts. Thus, our deep network jointly performs segmentation and prediction, while learning from 3D shapes with prescribed segments and mobilities.

\subsection{Motion inference for 3D objects} 
Works in computer graphics have also looked at the problem of motion inference for 3D objects. Mitra et al.~\shortcite{mitra2010} illustrate the motion of mechanical assemblies by predicting the probable motion of the mechanical parts and the entire assembly from the geometric arrangement of parts. Shao et al.~\shortcite{shao13} create animations of diagrams from concept sketches. For more general shapes, Pirk et al.~\shortcite{pirk2017} introduce interaction landscapes, a representation of the motion that an object undergoes while being used in a functional manner, e.g., a cup being used by a human to drink. This representation can then be used to classify motions into different types of interactions and also to predict the interactions that an object supports from a few seconds of its motion. 

Sharf et al.~\shortcite{sharf2013} capture the relative mobility of objects in a scene with a structure called a {\em mobility tree}. The tree is inferred from finding different instances of objects in different geometric configurations. Thus, while the method is able to infer the mobility of objects in a scene, it is limited by the assumption that multiple instances of the same objects appear in the input scenes. When given a single 3D object segmented into parts, Hu et al.~\shortcite{hu17} predict the likely motions that the object parts undergo, along with the mobility parameters, based on a model learned from a dataset of objects augmented with a small number of static motion states for each object. The model effectively links the geometry of an object to its possible motions. Yi et al.~\shortcite{yi18} predict the probable motion that the parts of an object undergo from two unsegmented instances of mobility-equivalent objects in different motion states. 

Differently from these works, our deep neural network RPM-Net predicts part motion from a 3D point cloud shape, without requiring a segmentation of the shape or multiple frames of the motion. We accomplish this by training a deep learning model to simultaneously segment the input shape and predict the motion of its parts.

Concurrently with our work, Wang et al.~\shortcite{wang2019shape2motion} also introduced an approach for mobility prediction without requiring a segmentation. Their approach first generates a set of proposals for moving parts and their motion attributes~\cite{wang2018}, which are jointly optimized to yield the final set of mobilities.
In contrast to their work, we break the analysis into motion prediction followed by mobility inference, which allows us to predict part motions even in instances when the motion cannot be described by a set of parameters. In addition, we obtain the mobility parameters directly by regression from the point cloud and predicted motion, rather than depending on an initial set of proposals.

\section{Overview of motion prediction} \label{sec:overview}

Our solution for motion and mobility prediction is enabled by two deep neural networks. The first network, RPM-Net, performs {\em motion hallucination}, predicting how the parts of an object can move. This network is the basic building block of our approach as it allows one to infer the moving parts of an object along with their motion. Moreover, if the user desires to summarize the predicted motion with a set of low-dimensional parameters, a second neural network, Mobility-Net, predicts the most likely transformation parameters that describe the motion predicted by the first network. In Section~\ref{sec:results}, we show that splitting the problem of mobility prediction into two separate networks allows us to obtain higher accuracy in mobility prediction, while also enabling the prediction of complex motions that cannot be easily described with a set of parameters, e.g., the opening of an umbrella.
As follows, we first describe the datasets that we use in our learning-based approach and evaluation, and then describe our two neural networks.

\section{Part mobility dataset} \label{sec:data}

\subsection{Dataset and mobility representation}
Since we use a learning-based approach for the prediction of object mobility, we require suitable training data. To create our training set, we were inspired by the data setup of Hu et al.~\shortcite{hu17} and obtained our dataset by extending their mobility dataset. Specifically, our dataset is a collection of shapes in upright orientation and segmented into parts. %
Each part is labeled either as a {\em moving part} or {\em
reference part}, where a shape has one reference part and one or more
moving parts.
For example, a bottle object could have a cap that twists (moving part) and handle for carrying (moving part) which both move in relation to a static liquid container (reference part). 

For each shape in the dataset, we take each possible pair composed of a moving and reference part, which we call a {\em mobility unit}, and associate a ground-truth mobility to this unit, specified as a set of parameters that describe the mobility of the moving part in relation to the reference part. The parameters are represented as a quadruple $(\tau, d, x, r)$, where $\tau$ is the transformation type (one of translation $T$, rotation $R$, or the translation-rotation combo $TR$), $d$ and $x$ are the direction and position of the transformation axis, and $r$ is the transformation range, stored as a start and end position (for translations) or angle (for rotations) relative to a local coordinate frame defined for each unit. %
Thus, the mobility information essentially encodes the possible motion of the parts without prescribing a specific speed for the transformation. %
The dataset contains $m=969$ objects, where 291 objects have more than one moving part, which results in 1,420 mobility units in total. Since our networks operate on point clouds, we sample the visible surface of the shapes to create point clouds with $N = 2,048$ points which we refer to as {\em complete scans}.

Moreover, one of the key advantages of our RPM-Net is that it is able to learn non-trivial motions beyond simple translation and rotation. To demonstrate this property, we also built a small dataset of shapes with non-trivial motions, which includes 24 umbrellas, 25  bows, and 21 balances.

\begin{figure}[!t]
    \centering
    \includegraphics[width=\linewidth]{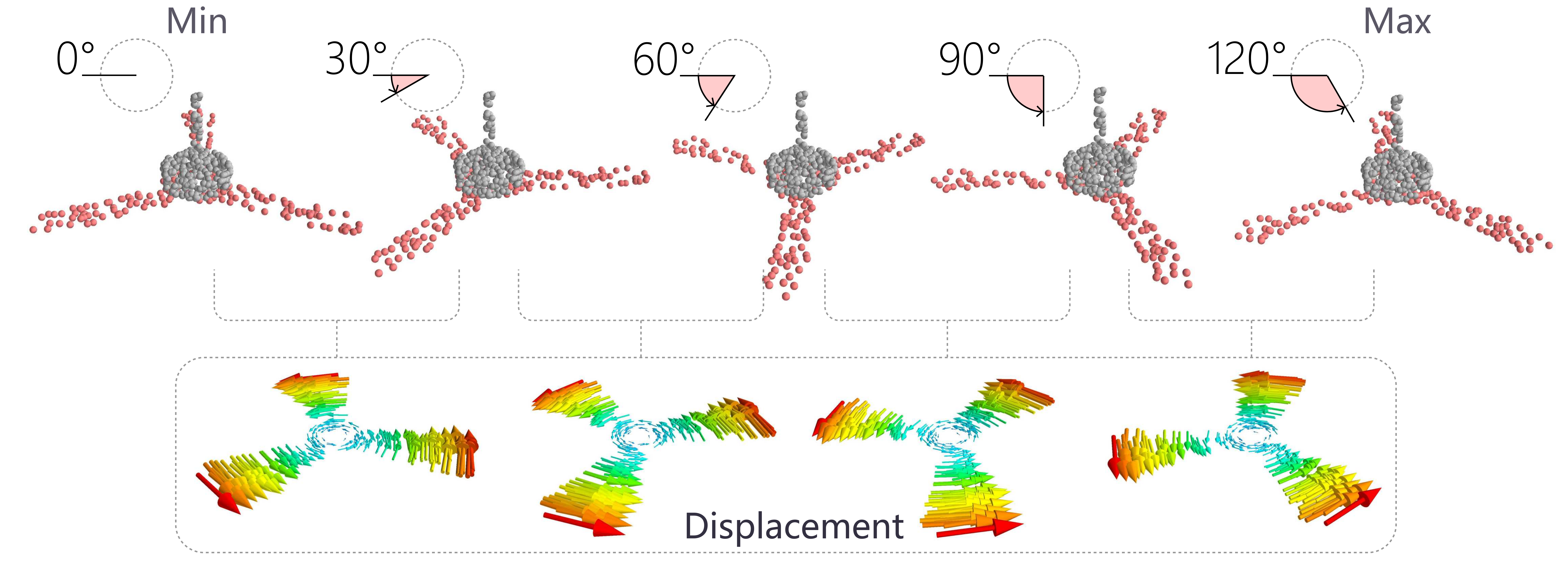}
\caption{Training data generation. For each mobility unit, we sample $n$ frames from the start to the end of the motion and compute the displacement field between each pair of adjacent frames. In this example, we see the sampling of the rotational motion of an electric fan, where the rotation angle range is defined to be $[0^{\circ}, 120^{\circ}]$ due to the rotational symmetry of the shape.}
\label{fig:frame_sample}
\end{figure}

\subsection{Training set}
To generate the training data for our approach, for the point cloud of each shape \rh{in our part mobility dataset}, we sample $n$ frames from the start to the end of the motion \rh{based on the motion parameters}, as illustrated in the top row of Figure~\ref{fig:frame_sample}. The selection of $n$ has the effect of adjusting the speed of the motions that will be learned and predicted.
For shapes with non-trivial motions, for which simple motion parameters are not available, we generate the motion sequences separately for each category. For umbrellas, we generate the motion sequence from the open to the closed state of each shape by moving points on the cover toward the handle, similar to the shrinking of a cone with linearly decreasing opening angles. For bows, the motion sequence captures the bending of the string and motion of the arrow, while the bow is kept rigid. Specifically, the arrow is translated along a horizontal axis, while the string is deformed from a bent to a straight state while keeping two points fixed.
For balances, we first generate a rotational motion for the bar and then set the motion of each pan to be the same as the motion of the point that connects the pan to the bar.

We design our approach so that, for each input frame at timestamp $t \in \{1,2, \dots, n\}$, it predicts an output sequence consisting of the $n-t$ frames after $t$. 
To be able to predict $n-t$ frames while ensuring that all the training data has the same dimensionality, and also to be able to infer when the motion stops, we duplicate the end frame a number of times at the end of each sequence, to make the length of the entire output sequence equal to $n$. In this manner, the relative motion of all the duplicated frames compared to their previous frames will be $0$, indicating that the end of the sequence has been reached.

Moreover, to generate ground truth data to guide our motion generation through an RNN, we also compute the displacement map between each pair of adjacent frames, which is simply defined as the difference between two consecutive point clouds along the motion, as shown in the bottom row of Figure~\ref{fig:frame_sample}.
To obtain the correspondence between moving points, we sample the point cloud on the static mesh first, and then apply the ground-truth motion to the point cloud to generate the individual frames. Thus, our training data is composed of $mn$ training instances, where one instance is a pair composed of the input frame and the output sequence of frames that should be predicted by the network.
Note that, for shapes with multiple moving parts, each generated frame and displacement map capture the simultaneous motion of the multiple parts. This enables us to predict the simultaneous motion for multiple moving parts of an object, as we will see in Section~\ref{sec:results}.

\subsection{Evaluation datasets of partial point clouds}
In addition to the set of complete point clouds described above, we also use synthetic and real datasets of partial point clouds in our evaluation. The synthetic dataset is generated by collecting virtual scans from random viewpoints of the 3D models in our complete dataset. More specifically, for each frame, we use the Kinect sensor model of Bohg et al.~\shortcite{bohg2014} and set random camera positions to generate the synthetic partial scans. Then, we transfer the ground-truth segmentation from the original mesh to the point clouds by comparing the distance from each point to the moving part and reference part in the mesh, and assigning the label of the closest part to the point. Moreover, the real dataset was obtained by scanning a variety of objects. We used a Microsoft Kinect v2 to scan big objects like baskets, and an Artec Eva to scan smaller objects such as a flip USB. For both datasets, since each partial scan can have a different number of points, we randomly subsample $N=2,048$ points from each point cloud while ensuring that each part is represented with at least 100 points.

\begin{figure*}[!t]
    \centering
    \includegraphics[width=0.98\textwidth]{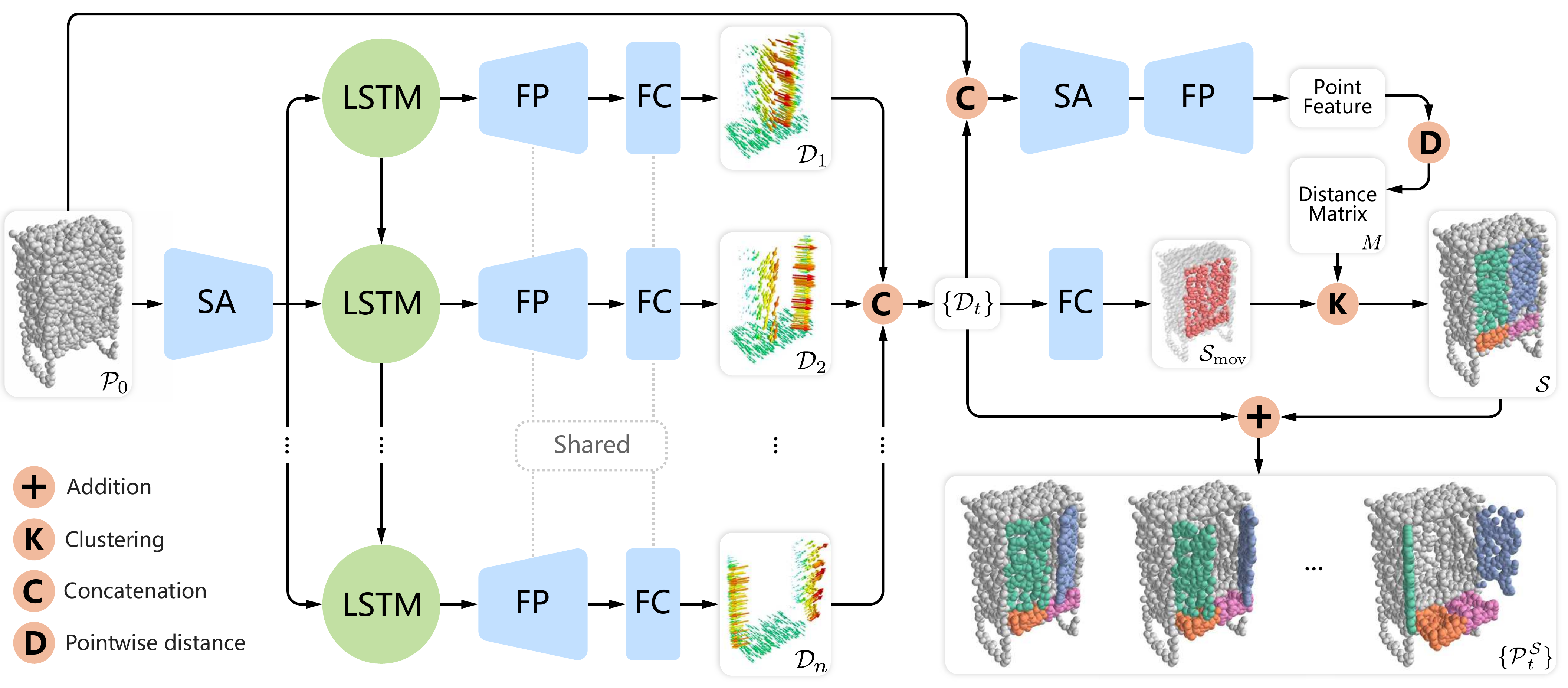}
\caption{The architecture of the motion hallucination network RPM-Net. Given an input point cloud $\mathcal{P}_0$, the network predicts displacement maps $\{ \mathcal{D}_t \}$ along with the segmentation $\mathcal{S}$ of the point cloud, which together provide the final segmented motion sequence $\{ \mathcal{P}_t^{\mathcal{S}}\}$. The network is composed of \rh{set abstractions} (SA), \rh{feature propagations} (FP), LSTM units, fully-connected layers (FC), and special operations denoted with the pink circles.}
\label{fig:gennet} 
\end{figure*}

\section{RPM-Net for motion hallucination}

\subsection{Input and output}
The input to the motion hallucination network RPM-Net is a point cloud $\mathcal{P}_0 \in \mathbb{R}^{N \times 3}$ that represents the sampling of a shape into a fixed number of points $N =2,048$. The output of the network is a set of displacement maps of the points, along with their segmentation. A single displacement map $\mathcal{D}_t \in \mathbb{R}^{N \times 3}$ represents the predicted motion of every point for a frame at time $t$, encoded as a spatial offset for each point. The complete network predicts a set of displacement maps $\{ \mathcal{D}_t \}$ for $n$ frames occurring after the input. For segmenting the point cloud, the network first classifies each point as belonging either to the reference part ($\mathcal{S}_{\text{ref}}$) or the moving part ($\mathcal{S}_{\text{mov}}$) of the shape. Then, the network further segments the moving part into isolated components $\{\mathcal{S}_{\text{mov}}^{i}\}_{i=1}^{C}$ with different sets of displacement fields. Note that objects may differ in the number of moving components $C$, which are determined automatically by clustering the points based on a pointwise distance matrix.

\subsection{Network architecture}
The core of RPM-Net is a Recurrent Neural Network (RNN) that predicts a displacement map. The RNN generates a displacement map that represents the motion of the moving parts of the object at a fixed time interval in the future. We employ this architecture since RNNs have demonstrated high accuracy in tasks related to the processing of sequential and time-series data. Specifically, RNNs model an internal state that enables them to dynamically adjust its processing according to temporal input, while at the same time learning features that capture long-term dependencies in sequential inputs. The specific RNN architecture that we use in our work employs Long Short-Term Memory (LSTM) components that form a type of temporal memory controlled with special gates connected to its inputs. 

A diagram of the full architecture of the motion hallucination network is shown in Figure~\ref{fig:gennet}. The first part of the network is composed of subnetworks that predict each of the $n$ frames of the motion, illustrated in the different rows of the figure, which provide a displacement map for each frame. In each of these subnetworks, we use the PointNet++~\cite{qi17} encoder to create an $\mathbb{R}^{2048}$ feature vector for the input point cloud $\mathcal{P}_0$, which represents the points in a manner that is invariant to the order of points given in the input. This feature vector is then fed to an LSTM that learns the relationships between the features and the temporal displacement of the points. 
More specifically, we set the initial state of the LSTM to a vector of zeros. After feeding the feature vector and zero vector into the LSTM, the unit returns the next state and another output that represents the displacement map. A PointNet++ decoder and sets of fully-connected layers then decode the output of the LSTM into a displacement map. Next, we provide the next state along with the feature vector as input to the LSTM again, to obtain the subsequent displacement map. We repeat this procedure $t$ times to obtain all the displacement maps $\mathcal{D}_t$.

In the second part of the network, the displacement maps $\{\mathcal{D}_t\}$ are concatenated and passed with the input point cloud $\mathcal{P}_0$ to a segmentation module that provides a motion-based segmentation $\mathcal{S} = \left\{\mathcal{S}_{\text{ref}}, \mathcal{S}_{\text{mov}}^1, \dots, \mathcal{S}_{\text{mov}}^C\right\}$. 
More specifically, $\{\mathcal{D}_t\}$  is first passed to additional fully-connected layers that predict the object-level segmentation of the point cloud which segments the shape into a reference part $\mathcal{S}_{\text{ref}}$ and a moving part $\mathcal{S}_{\text{mov}}$. To further segment $\mathcal{S}_{\text{mov}}$ into multiple moving parts, we cluster $\mathcal{S}_{\text{mov}}$ into separate groups according to point set features. Specifically, $\{\mathcal{D}_t\}$ is concatenated with the input cloud $\mathcal{P}_0$ and then fed to a PointNet++~\cite{qi17} encoder-decoder that extracts a set of point features. Next, a pointwise distance matrix $M$ is obtained by computing the Euclidean distance between corresponding point features. %
We derive a submatrix $M_\text{mov}$ for the points in $\mathcal{S}_{\text{mov}}$ and cluster it with DBSCAN~\cite{ester1996density}, to separate the points in the moving part into $C$ groups corresponding to $C$ moving components, where $C$ is automatically determined by the algorithm. 

Note that our approach to obtain a variable number of segments from a pointwise distance matrix is inspired by the similarity group proposal network of Wang et al.~\shortcite{wang2018}. However, instead of considering the rows of the pointwise matrix as different group proposals and then employing a non-maximum suppression step to generate the final segmentation, we consider $M_\text{mov}$ as a metric matrix and apply a clustering algorithm directly to the matrix to produce the segments. In this way, all the pairwise relationships between points are considered when forming the clusters.

\subsection{Network training and loss functions}
To train this multi-output network, we design suitable loss functions to
account for each type of output. The loss for the motion hallucination network is defined as:
	\begin{equation}
	L(\{\mathcal{D}_t\}, \mathcal{S}) =
	\rh{\frac{1}{n}}\left(\sum_{t=1}^{n}{L_{\text{rec}}(\mathcal{D}_t)}\right) +
	L_{\text{mot}}(\{\mathcal{D}_t\}) + 
	L_{\text{seg}}(\mathcal{S}) ,
	\label{eq:loss_h}
	\end{equation}	
	where $L_{\text{rec}}$ is the reconstruction loss, $L_{\text{mot}}$ is the motion loss, and  $L_{\text{seg}}$ is the segmentation loss. The first two terms ensure that an accurate displacement map is generated for each frame in the sequence, and the segmentation loss $L_{\text{seg}}$ ensures the correct separation of the moving and reference parts. We now discuss each term in detail.

\subsubsection{Reconstruction loss}
The loss $L_{\text{rec}}$ quantifies the reconstruction quality of both
the point cloud and displacement maps. Specifically, it measures the
deviation of the point cloud from its original geometry after motion with a term $L_{\text{geom}}$, and the deviation of the displacement maps from the ground truth with a term $L_{\text{disp}}$:
\begin{equation}
L_{\text{rec}}(\mathcal{D}_t) = L_{\text{geom}}(\mathcal{P}_t) + L_{\text{disp}}(\mathcal{D}_t),
\end{equation}
where $\mathcal{P}_t = \mathcal{P} _{t-1}+\mathcal{D}_t$ is the point cloud
after displacement.

The loss for the geometry is given by:
\begin{equation}
L_{\text{geom}}(\mathcal{P}_t) = \rh{\omega_{\text{ref}}}\, L_{\text{ref}}(\mathcal{P}_t) + \rh{\omega_{\text{mov}}}\, L_{\text{mov}}(\mathcal{P}_t),
\end{equation}
where $L_{\text{ref}}(\mathcal{P}_t)$ is the loss of the reference part, $L_{\text{mov}}(\mathcal{P}_t)$ is a loss that considers all the moving parts together,  \rh{and the corresponding weights are set as $\omega_{\text{ref}} = 10$ and $\omega_{\text{mov}} = 5$.}

To measure the geometric distortion, we use the ground-truth segmentation to split $\mathcal{P}_0$ into reference and moving parts. For each time-frame, the reference part should be kept static while the geometry of the moving parts should only be transformed rigidly. Thus, 
to measure the distortion of the reference part, we simply compare its point positions in the original and predicted point clouds:
\begin{equation}
L_{\text{ref}}(\mathcal{P}_t) = \sum_{p \in \mathcal{P}_{\text{ref}}^t} \|p - p^{ \text{gt}}\|, 
\end{equation}
where $p^{\text{gt}}$ is the original position of point $p$. To account for the moving parts, we make use of a loss introduced by Yin et al~\shortcite{yin18} to measure the geometric difference between two point sets:
\begin{equation}
L_{\text{mov}}(\mathcal{P}_t) = L_{\text{shape}}(\mathcal{P}_{\text{mov}}^{t}, \mathcal{P}_{\text{mov}}^{t,\text{gt}}) + L_{\text{density}}(\mathcal{P}_{\text{mov}}^{t}, \mathcal{P}_{\text{mov}}^{t,\text{gt}}),
\end{equation}
where the $L_{\text{shape}}$ term penalizes points that do not match with the target shape, while $L_{\text{density}}$ measures the discrepancy of the local point density between two corresponding point sets~\cite{yin18}. In practice, these terms are computed by finding the closest points from the original to the transformed point cloud.

To compute the displacement loss $L_{\text{disp}}(\mathcal{D}_t)$, we simply compute the difference between the ground truth and predicted displacement for each point:
	\begin{equation}
	L_{\text{disp}}(\mathcal{D}_t) = \sum_{p \in \mathcal{P}_{\text{mov}}} \left\|\mathcal{D}_{t}(p) - \mathcal{D}_{t}^{ \text{gt}}(p)\right\|.
	\end{equation}

\subsubsection{Motion loss}
The motion loss $L_{\text{mot}}(\{\mathcal{D}_t\})$ is used to ensure the smoothness of the hallucinated motion. More specifically, we constrain the displacement of a point to be consistent across adjacent frames of the motion, i.e., $\|\mathcal{D}_{t_i}(p)\|_2 = \|\mathcal{D}_{t_{i+1}}(p)\|_2$, $\forall p \in \mathcal{P}_{\text{mov}}$, which we capture with a variance-based loss:
	\begin{equation}
	L_{\text{mot}}(\{\mathcal{D}_t\}) =
	\sum_{p \in \mathcal{P}_{\text{mov}}} \sigma^2\left(\left\{ \left\|\mathcal{D}_t(p)\right\|_2 \right\}_{t=0}^{n}\right),
	\end{equation}
\rh{where $\sigma^2(\ldots)$ is the variance of a set of observations.}

\subsubsection{Segmentation loss}
Since the number of moving components changes for different input shapes, we define the segmentation loss over the intermediate results, i.e., the object-level segmentation given by $\{\mathcal{S}_{\text{ref}}, \mathcal{S}_{\text{mov}}\}$ and the pointwise distance matrix of the moving part $M_{\text{mov}}$, which is used to obtain the finer segmentation of $\mathcal{S}_{\text{mov}}$. Thus, the segmentation loss is defined as:
\begin{equation}
	L_{\text{seg}}(\mathcal{S}) = 
\rh{	\omega_{\text{seg}}^{\text{obj}}\,  }L_{\text{seg}}^{\text{obj}}(\mathcal{S}_{\text{ref}},  \mathcal{S}_{\text{mov}}) +
\rh{  \omega_{\text{seg}}^{\text{mov}}\, }L_{\text{seg}}^{\text{mov}} (M_{\text{mov}}),
        \label{eq:seg_loss}
\end{equation}	
where $L_{\text{seg}}^{\text{obj}}$ is the object-level segmentation loss defined as the softmax cross entropy between the predicted segmentation and the ground-truth segmentation, and $L_{\text{seg}}^{\text{mov}}$ is the loss for the finer segmentation of the moving part, defined as:
\begin{equation}
L_{\text{seg}}^{\text{mov}} (M_{\text{mov}}) = \sum_{i=1}^{N_{\text{mov}}}\sum_{j=1}^{N_{\text{mov}}} l(m_i,m_j),
\end{equation}
where $N_\text{mov}$ is the number of points in the moving part and $l(m_i,m_j)$ is defined as:
\begin{equation}
l(m_i,m_j) = 
\begin{cases}
M_{\text{mov}}(i,j), 	& \text{if}\ M_{\text{mov}}^{\text{gt}}(i, j) = 0, \\
\max(0, K-M_{\text{mov}}(i, j)),   & \text{if}\ M_{\text{mov}}^{\text{gt}}(i,j) = 1,
\end{cases}
\end{equation}
with $M_{\text{mov}}^{\text{gt}}$ being the ground truth distance matrix, where an entry $(i,j)$ has the value 0 if points $i$ and $j$ belong to the same moving part, and 1 otherwise. $K$ is a constant margin which we set to the default value of $80$. 
\rh{The corresponding weights are set as $\omega_{\text{seg}}^{\text{obj}} = 2$ and $\omega_{\text{seg}}^{\text{mov}} = 0.2$.}

\section{Mobility-Net for parameter prediction}

\subsection{Input and output}
For the prediction of mobility parameters, the input to Mobility-Net is the point cloud $\mathcal{P}_0 $ together with the set of displacement maps $\{ \mathcal{D}_t^{i} \}$ predicted by RPM-Net for the $i$-th moving component $\mathcal{S}_{mov}^{i}$, for which we would like to infer the mobility parameters. The output is a set of mobility parameters $\mathcal{M}_i$ which describe at a high level the mobility of the component through all the frames. The parameters are encoded as a tuple $\mathcal{M}_i = (\tau_i, d_i, x_i)$, where $\tau_i$ is the transformation type, and $(d_i, x_i)$ are the direction and position of the transformation axis. Note that our method does not estimate the remaining mobility parameter, the transformation range $r$, since we can derive the range from the position of the moving part in the start and end frames.

\begin{figure}[!t]
    \centering
    \includegraphics[width=0.94\linewidth]{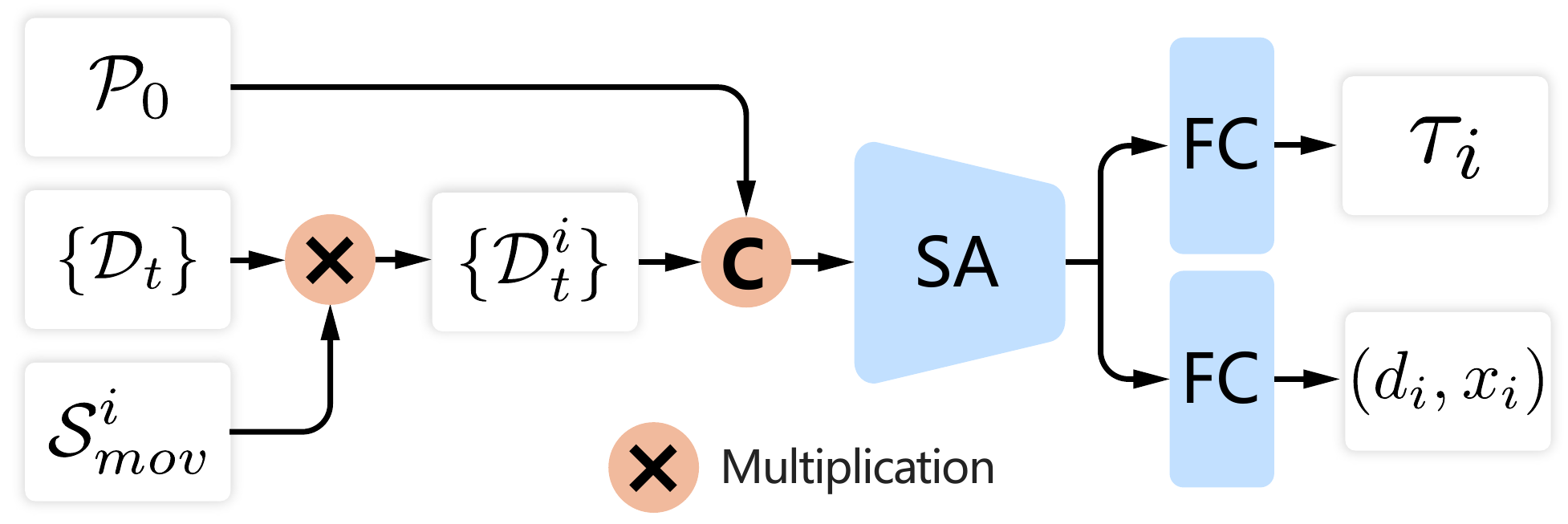}
\caption{The architecture of the mobility prediction network Mobility-Net. For each segmented moving component $\mathcal{S}_{mov}^{i}$, the network takes the point cloud $\mathcal{P}_0$ and the corresponding generated displacement maps $\{ \mathcal{D}_t^i\}$ as input to predict the high-level mobility parameters $(\tau_i, d_i, x_i)$. The network is composed of an encoder (SA) and fully-connected layers (FC).}
\label{fig:prednet} 
\end{figure}

\subsection{Network architecture}
The architecture of Mobility-Net is shown in Figure~\ref{fig:prednet}. For each moving component $\mathcal{S}_{mov}^{i}$, we take its corresponding set of displacement maps $\{ \mathcal{D}_t^{i} \}$ and concatenate it with the point cloud $\mathcal{P}_0$ to obtain an input vector of dimension $\mathbb{R}^{N \times (n+1) \times 3}$. This vector is passed to a PointNet++~\cite{qi17} encoder that extracts features of the input, which are then passed to two additional fully-connected layers, which provide the corresponding set of mobility parameters $\mathcal{M}_i = (\tau_i, d_i, x_i)$. Since the transformation type $\tau_i$ is an integer, while the other two mobility parameters axis direction $d_i$ and position $x_i$ are real values, we use two separate subnetworks for their prediction.

\subsection{Loss function}
The loss for the mobility prediction network ensures the correctness of the transformation type as well as the position and direction of the transformation axis, for each segmented moving component $\mathcal{S}_{mov}^{i}$. 
The loss of the mobility parameters $\mathcal{M}_i$ for the $i$-th moving part is defined as:
\begin{equation}
L_{\text{mob}}(\mathcal{M}_i) = H(\tau_i, \tau_i^{\text{gt}}) + \|d_i - d_i^{\text{gt}}\|_2 + \|x_i - x_i^{\text{gt}}\|_2,
\label{eq:mob_loss}
\end{equation}
where $\tau_i$, $d_i$, and $x_i$, are the predicted transformation type, motion axis direction, and position, respectively, while $\tau_i^{\text{gt}}$, $d_i^{\text{gt}}$, and $x_i^{\text{gt}}$, are the ground-truth values, and $H$ is the cross-entropy. 

\section{Results and evaluation} \label{sec:results}

In this section, we show results of motion hallucination and mobility prediction obtained with our networks RPM-Net and Mobility-Net, and evaluate different components of the approach.
As described in Section~\ref{sec:data}, for our experiments, we use
three datasets, which include one dataset of complete points clouds that
we split into a 90/10 ratio for training/testing, and two additional
evaluation sets composed of synthetic and real partial scans.
The three datasets consist of shapes with single and multiple moving parts. Since for shapes with a single moving part we only need to generate the motion sequences and do not need to further segment the moving parts, we show results for shapes with single and multiple moving parts separately.

We first present a set of qualitative results to demonstrate the capabilities of our method. A quantitative evaluation and ablation studies of the method are shown in Sections~\ref{sec:quantitative} and~\ref{sec:ablation}.

\subsection{Qualitative results} 
\label{seq:qualitative}

We show results for synthetic shapes with single and multiple moving parts, and also results of motion hallucination for shapes with non-trivial motion and hierarchical motion. To demonstrate the generality our method, we also apply our networks to real scans. 

\begin{figure*}[!t]
    \centering
    \includegraphics[width=\textwidth]{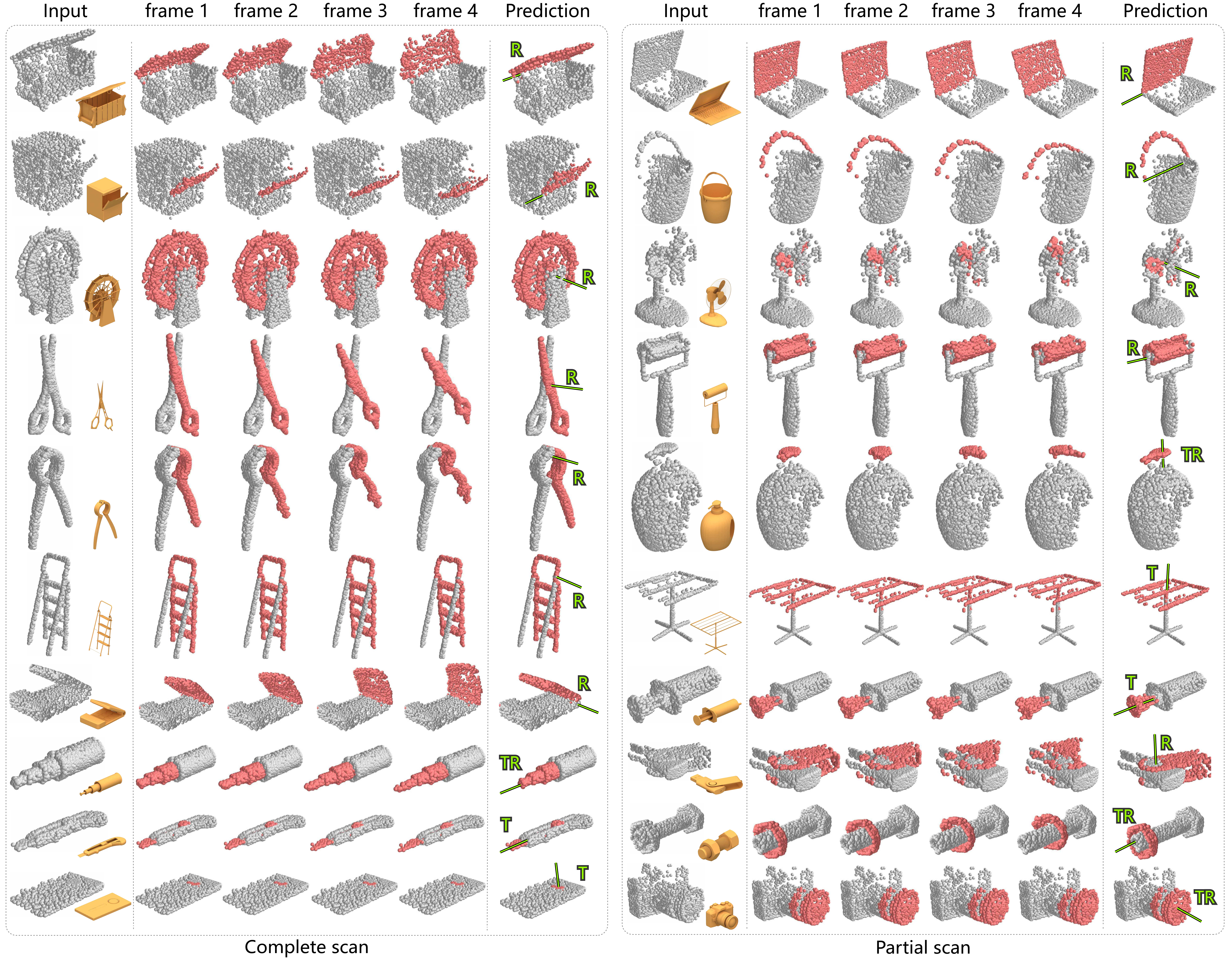}
\caption{Motion prediction results on shapes with a single moving part. We observe how our method can be applied to a variety of shapes with diverse mobilities, including both complete point clouds and partial scans. For each input cloud, we show the first four frames of the predicted motion, along with the predicted transformation axis drawn as a green line, and moving and reference parts colored red and gray, respectively. We observe how RPM-Net can predict the correct motion sequences for different inputs and estimate the corresponding part mobility parameters.}
\label{fig:gallery}
\end{figure*}

\subsubsection{Results on shapes with a single moving part}
Figure~\ref{fig:gallery} shows visual examples of mobility prediction on test units for both complete and partial scans. For each example, we show the first four predicted frames for each input point cloud with the predicted transformation axis drawn as a green line on the segmented input point cloud. The reference part is colored in gray while the moving part is red. We observe how RPM-Net can generate correct motion sequences for different object categories with different motion types and predict the corresponding mobility of the moving parts. 

For example, our method predicts rotational motions accurately on shapes with different axis directions and position. This includes the prediction of both horizontal and vertical axis directions, such as the flip phone in the seventh row (left) versus the twist flash drive in the eighth row (right), and the correct prediction of axis positions, whether they are close to the center of the moving part or to its side, as in the case of the scissors in the fourth row (left) versus the pliers in the fifth row (left). Our method is able to predict the correct motion and mobility parameters even for shapes where the moving part is partially occluded by the reference part, e.g., the fan on the third row (right), where a portion of the blade is occluded by its protective case. 

We can also see that, for translational motions, for example, the cutter in the ninth row (left), RPM-Net is able to predict the correct direction along which the blade opens by translation, although the data only presents part of the blade while the reference part that wraps around the blade is much more pronounced than the blade itself.  Moreover, we can see that the small button moves together with the blade during the motion.  We see a similar result of correct translation detection for the syringe in the seventh row (right).

Our method can also correctly predict motions that involve a combination of translations and rotations. For example, the head of the bottle in the fifth row (right) is correctly rotated and translated upwards. For other examples with more symmetric moving parts, even when the rotational motion cannot be seen clearly by a human from the static models, our network can generate the rotational motion and correctly predict the motion type as shown for the telescope in the eighth row (left) and the nut in the ninth row (right). 
Moreover, we can see from the camera shown in the last row (right) that, for input point clouds that are already close to the end frame, our method learned to stop generating new frames after finding the end state of the motion, which shows that our method is able to infer the motion range.

Please see the accompanying video for a dynamic visualization of these predicted motions.

\begin{figure}[!t]
    \centering
    \includegraphics[width=0.99\linewidth]{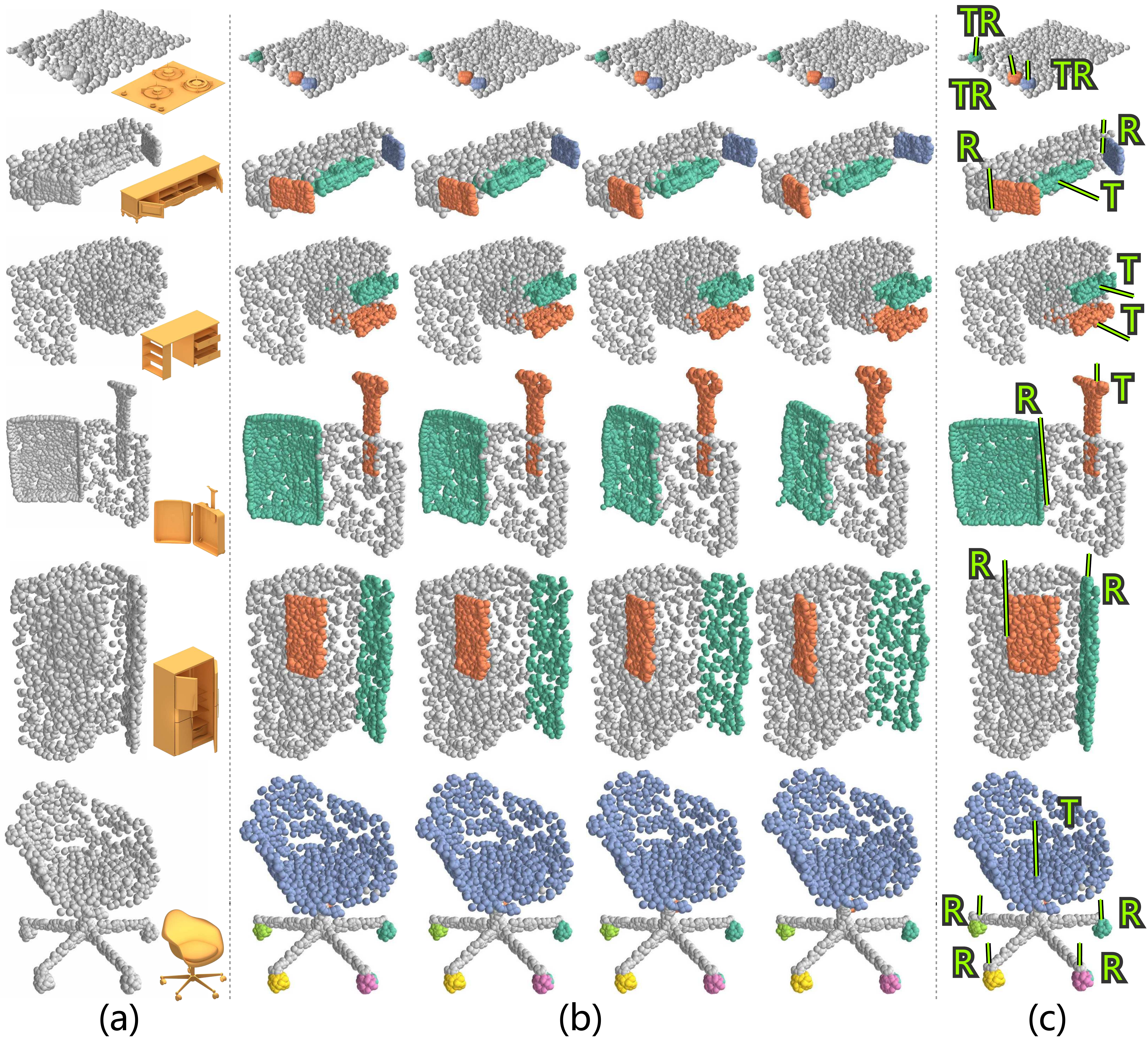}
\caption{Motion prediction results on shapes with multiple moving parts. (a)~Input point cloud, (b) Motion hallucinated with our method (four frames), (c) Segmentation with predicted motion axis for each moving part.}
\label{fig:multi_mob}
\end{figure}

\subsubsection{Results on shapes with multiple moving parts}
Given an object with multiple part mobilities, RPM-Net can simultaneously predict the motion of all the moving parts. Figure~\ref{fig:multi_mob} shows examples of simultaneous motion prediction, where we show four consecutive frames of the predicted motion along with the predicted segmentation and transformation axis, where each moving part is drawn with a different color, while the reference part is shown in gray. The transformation axis is drawn as a green line over each moving part on the segmented point cloud shown in the last column.
 
We see that our method can correctly segment moving parts in several different configurations. For the gas stove shown in the first row, our method successfully segments the three switches and correctly predicts their corresponding mobility parameters. For the TV bench shown in the second row, our method is able to recognize different types of motion and segment the two doors and the drawer into different moving parts. The other examples show similar results.

\subsubsection{Results on shapes with non-trivial motion}
\begin{figure}[!t]
    \centering
    \includegraphics[width=0.99\linewidth]{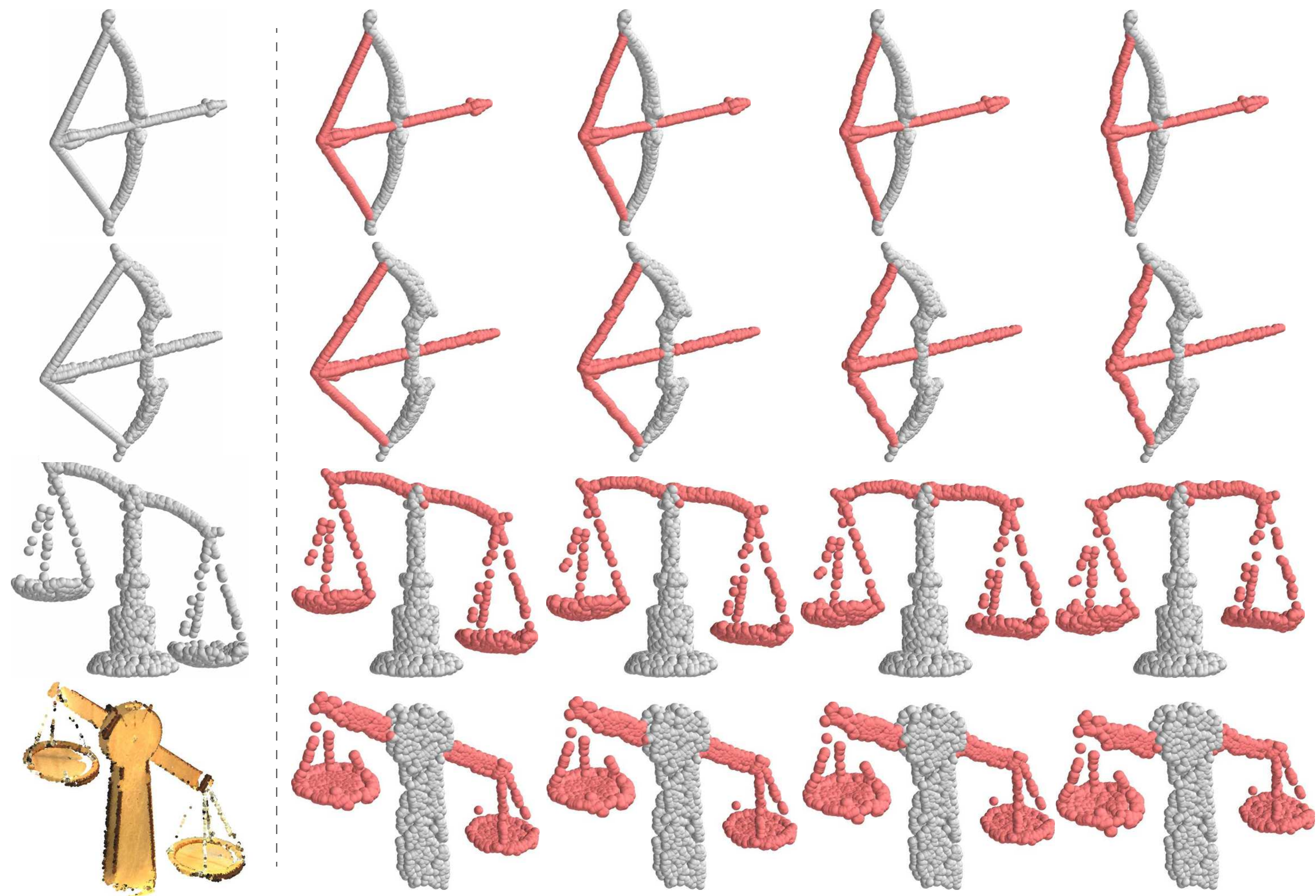}
\caption{Non-trivial motion prediction results: motion hallucinated for \ok{two bows (both synthetic) and two balances (one synthetic and one a real scan)}, which cannot be described with a simple transformation.}
\label{fig:nontrivial_motion}
\end{figure}

Given that our motion hallucination network RPM-Net is able to learn complex
motions independently of whether these motions can be described by mobility parameters and estimated by Mobility-Net, our method can also perform
hallucination for shapes with non-trivial motions, i.e., where the motion cannot be simply classified into translation, rotation, or the translation-rotation combination. As a consequence, non-trivial motions cannot be represented with a set of low-dimensional mobility parameters $\mathcal{M}$. 
Figure~\ref{fig:nontrivial_motion} shows \rh{four} examples of non-trivial motion hallucinated by RPM-Net, where our method correctly predicted motion sequences for \rh{two bows and two balances}. In the bow example, the motion captures the displacement of the arrow as well as the shrinking of the bow's string. The motion of the balance includes the swinging of the pans.
\ok{Note that, even though the shapes of the two bows and two balances are quite different, our method is able to generate correct motion sequences for these shapes, which include synthetic inputs and real scans, demonstrating the robustness of our method. Moreover, } 
most previous works cannot deal with such types of motion~\cite{hu17, yi18, wang2019shape2motion}, demonstrating the generality of our motion hallucination network. Note that, to obtain these results with more complex motions, we trained our network separately for different object categories that have different motions.

\begin{figure}[!h]
    \centering
    \includegraphics[width=0.9\linewidth]{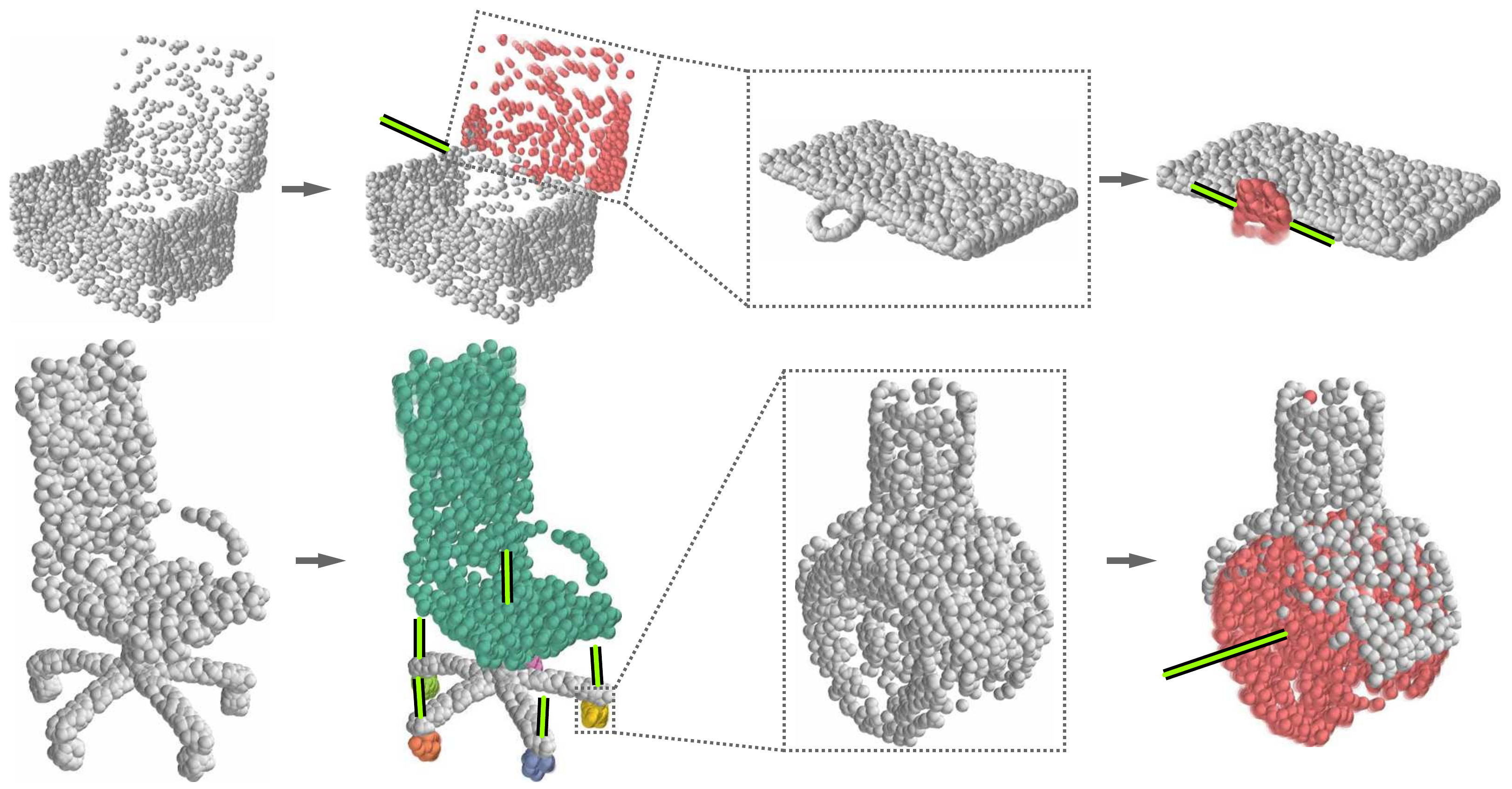}
\caption{Results of hierarchical motion prediction, where we detect mobility at two levels of a hierarchy by applying RPM-Net recursively.}
\label{fig:multi_level}
\end{figure}

\subsubsection{Results on shapes with hierarchical motions}
We can apply RPM-Net recursively to predict the mobility of parts organized hierarchically (Figure~\ref{fig:multi_level}). In the first example, we first detect the rotational opening of the chest's cover, followed by the detection of the motion of the cover's handle. In the second example, we first detect the vertical rotation of the seat and wheels of the chair, followed by the detection of the horizontal rotation of the wheels at a finer scale.

\begin{figure}[!t]
    \centering
    \includegraphics[width=0.99\linewidth]{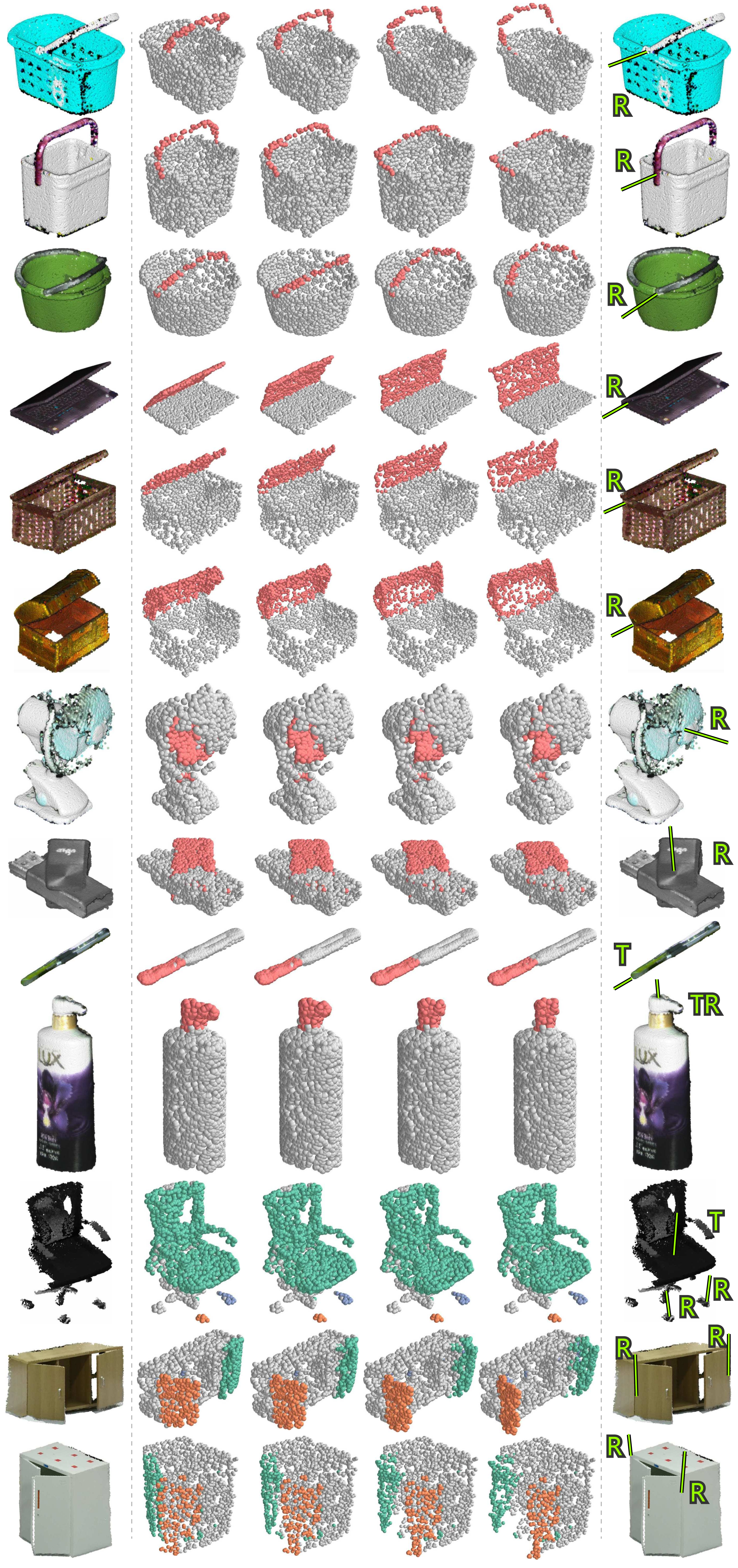}
\caption{Motion prediction results on real scans, where the moving
parts are shown with different colors and the reference part is colored gray.} 
\label{fig:real_scan}
\end{figure}

\subsubsection{Results on real scans}
Figure~\ref{fig:real_scan} shows that we can also apply our RPM-Net on real scans of objects. The last three rows show objects scanned with a Kinect v2, while the other smaller objects were scanned with an Artec Eva scanner.
We can see that our method can handle shapes with different complexity, different numbers of moving parts, and different types of motions. Although the baskets shown in the first three rows have quite distinct shapes, our method can predict a meaningful motion for their handles and also predict the correct motion type and axis position for the motion. Accurate predictions can also be observed for the three following objects which are of different categories including a laptop and two boxes. Moreover, rotational motions can also be correctly predicted for shapes with occluded parts, such as the fan shown in the seventh row with an incomplete point cloud, and the flash drive shown in the eighth row which has a different axis direction. Our method can also predict translations and translation-rotation combo motions, and the corresponding example results are presented in the ninth and tenth rows, respectively. For shapes with more complex structures and multiple moving parts, even if the scans are incomplete, our method can still predict reasonable motions for most of the parts scanned, as shown in the last three rows.

\begin{figure}[!t]
    \centering
    \includegraphics[width=0.94\linewidth]{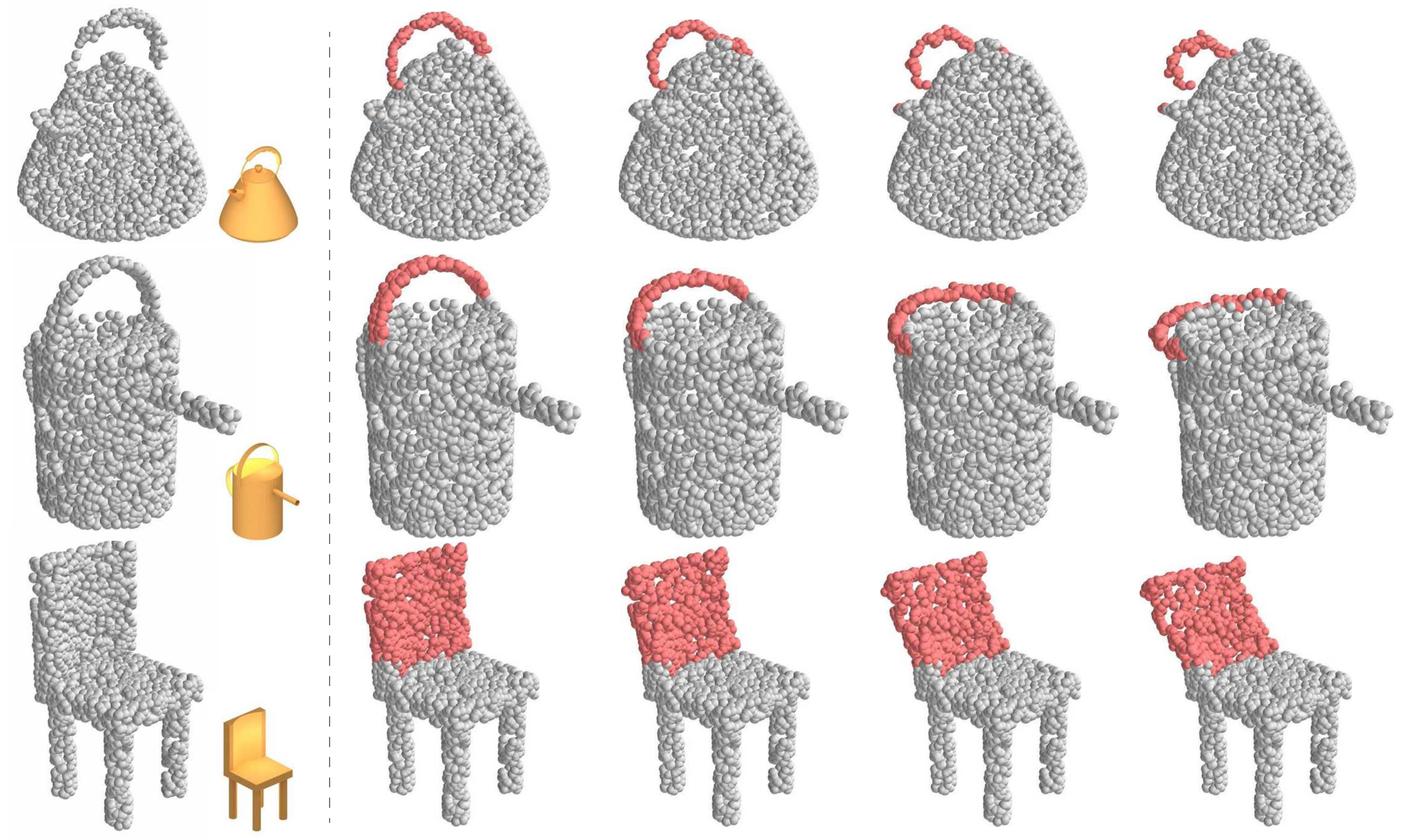}
\caption{Motion prediction results for out-of-distribution objects, i.e., objects from categories that do not exist in our dataset. We show results for 
a kettle, a sprinkler, and a four-legged chair.}
\label{fig:unseen}
\end{figure}

\subsubsection{Results on out-of-distribution objects}
To further evaluate the generality of our method, we apply our method to shapes that do not belong to any category in our dataset. A few example results are shown in Figure~\ref{fig:unseen}. We see that, for the kettle and sprinkler in the first two rows, our method can successfully predict the correct motion of the handles, although the geometry of the bodies of these shapes is quite different from shapes in our training set. For the four-legged chair, we predict the folding motion of its back. Ultimately, what is critical for obtaining successful results such as these is not the coverage of the categories themselves, but how much similarity exists between the test and training shapes.

\subsection{Quantitative evaluation}
\label{sec:quantitative}
We perform a quantitative evaluation of mobilities predicted by RPM-Net + Mobility-Net for the test set by measuring the error in mobility parameters and segmentation, for which we have a ground truth available. Specifically, for each test moving part, we compute the error of the predicted transformation axis $\mathcal{M} = (d, x)$ compared to the ground-truth axis $\mathcal{M}^{\text{gt}} = (d^{\text{gt}}, x^{\text{gt}})$ with two measures. The first measure accounts for the error in the predicted axis direction:
\begin{equation}
E_{\text{angle}} = \text{arccos}(\big|\text{dot} (d/\|d\|_2, d^{\text{gt}}/\|d^{\text{gt}}\|_2)\big|),
\end{equation}
which is simply the angle of deviation between the predicted and ground-truth axes, in the range $[0, \pi/2]$. The second measure computes the error for the position of the axis:
\begin{equation}
E_{\text{dist}} = \min(\|x^{\text{gt}} - \pi(x^{\text{gt}})\|_2, 1),
\end{equation}
where $\pi(x^{\text{gt}})$ projects point $x^{\text{gt}}$ from ground-truth transformation axis onto the predicted transformation axis determined by $\mathcal{M} = (d,x)$. Since all the shapes are normalized into a unit box, we truncate the largest distance to 1. Note that translations do not have an axis position defined. Thus, we only compute the axis direction error for translations.
The transformation type error $E_{\text{type}}$ is set to be 1 when the classification is incorrect and 0 otherwise. 
For shapes with multiple moving parts, we compute the errors for all the parts and get the average error. To compute the error for each ground-truth part, we find the predicted part with maximal IoU.

To measure the segmentation accuracy, we use the Average Precision (AP) metric defined as:
\begin{equation}
    \text{AP}=\Sigma_{k=1}^{10}(R_{k}-R_{k+1})P_k,
\end{equation}
where the pair $(P_k, R_k) $ is the precision-recall pair computed using the threshold of index $(11 -k)$ from the set $[0.5:0.05:0.95]$.
We set the $11$-th pair $(P_{11}, R_{11})$ as $(1, 0)$. Then, we define the segmentation error as $E_{\text{seg}} = 1 - \text{AP}$. Note that AP defined in this manner is the primary evaluation measure of the COCO challenge~\cite{COCO}.

When comparing to the baseline BaseNet, explained further down, we consider only shapes with a single moving part, since BaseNet does not handle multiple moving parts. We compute the mean for each error measure for two test sets, complete point clouds and synthetic partial scans, an report the result in Table~\ref{tab:comp_basenet}. For the whole dataset of shapes with one or more moving parts, we report the averages on the set of complete shapes in the last row of Table~\ref{tab:comp_shape2motion}. We observe in these two tables that all the errors are relatively low, confirming the trend seen in the qualitative results and indicating that the accuracy of the predicted mobility is high throughout the datasets. Moreover, for shapes with a single moving part, our method reaches comparable performance for both complete and partial scans, showing the robustness of our method to incomplete data.

\begin{figure}[!t]
    \centering
    \includegraphics[width=0.75\linewidth]{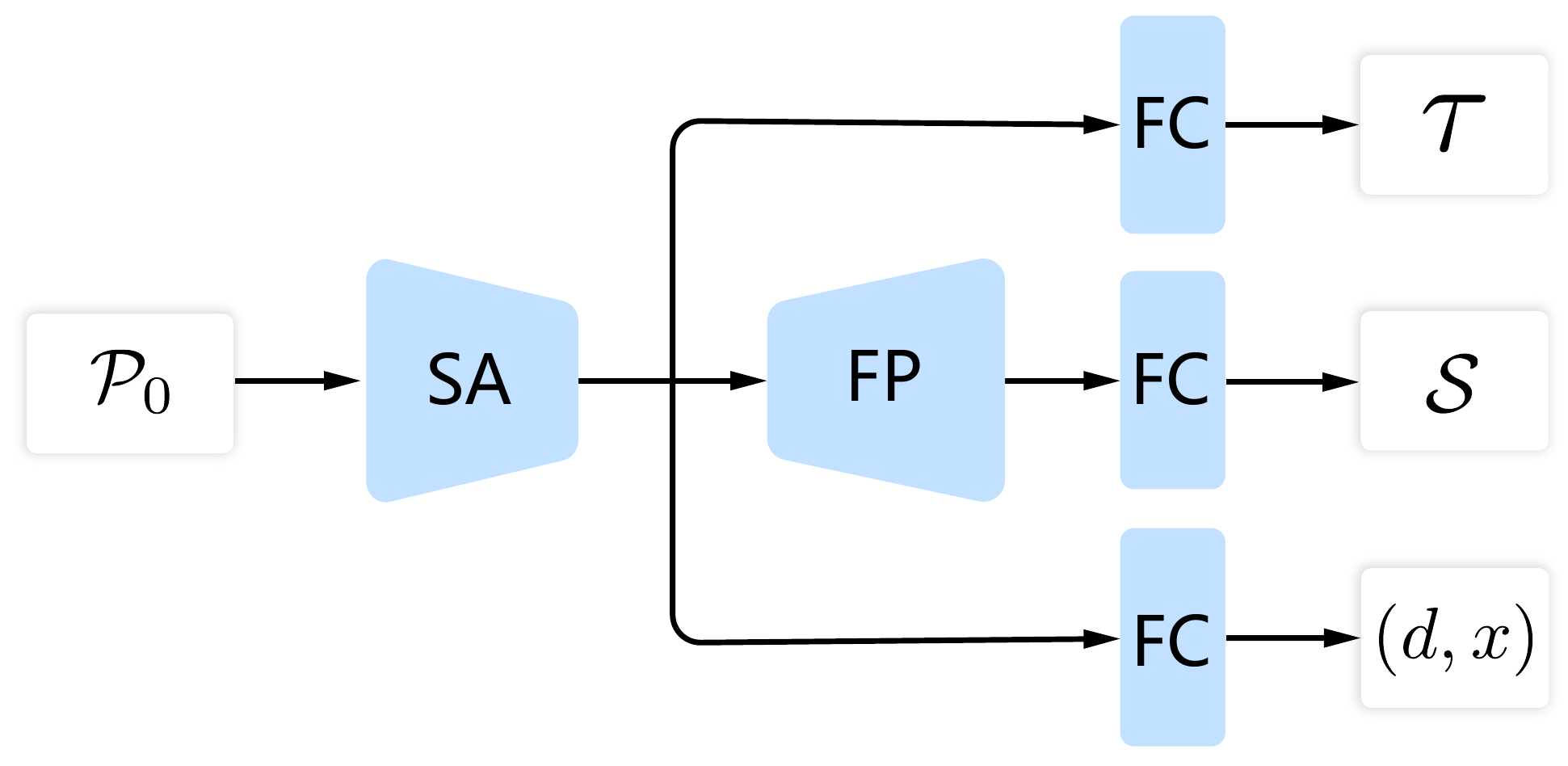}
\caption{The architecture of the baseline prediction network ``BaseNet''. }
\label{fig:basenet}
\end{figure}

\subsubsection{Comparison to BaseNet.}
To show the advantage of using RPM-Net, which generates the displacement maps before predicting all the mobility parameters, we compare our network to a baseline, which we call ``BaseNet''. BaseNet takes the point cloud $\mathcal{P}_0$ as input and estimates the segmentation $\mathcal{S}$ and mobility parameters $\mathcal{M}$ directly with a standard network architecture. The network is composed of an encoder/decoder pair and fully-connected layers, illustrated in Figure~\ref{fig:basenet}. The loss function for BaseNet is: 
\begin{equation}
L(\mathcal{S}, \mathcal{M}) =
L_{\text{seg}}^{\text{obj}}(\mathcal{S}) + 
L_{\text{mob}}(\mathcal{M}),
\end{equation}
which uses our losses defined in Equations~\ref{eq:seg_loss} and~\ref{eq:mob_loss}.

\begin{table}[!t]%
	\caption{Errors in motion-based segmentation and mobility prediction for our method and BaseNet.}
	\label{tab:comp_basenet}
	\begin{minipage}{\columnwidth}
		\begin{center}
		\begin{tabular}{l||l||l|l|l|l}
			\hline
			\textbf{Data type} & \textbf{Method} &  $E_{\text{seg}}$ & $E_{\text{angle}}$&  $E_{\text{dist}}$  &  $E_{\text{type}}$ \\ \hline
			\multirow{2}{*}{Complete scans} & BaseNet & 0.182 & 0.260 &  0.301 &  0.074  \\ \cline{2-6} 
			& \textbf{Ours} & \textbf{0.161}    & \textbf{0.126} & \textbf{0.166}  & \textbf{0.014}  \\ \hline
			\multirow{2}{*}{Partial scans}  & BaseNet &  0.235 & 0.319 &  0.289  & 0.086 \\ \cline{2-6} 
			& \textbf{Ours}   & \textbf{0.199}  & \textbf{0.20} & \textbf{0.198}  & \textbf{0.020} \\ \hline
		\end{tabular}
		\end{center}
	\end{minipage}
\end{table}%

Note that, since BaseNet cannot handle shapes with multiple moving parts as different shapes may have different numbers of moving parts, we perform our comparison only on shapes with a single moving part. This is also the reason why the loss for BaseNet does not involve a term for the moving part segmentation quality. Table~\ref{tab:comp_basenet} shows the comparison between RPM-Net + Mobility-Net and BaseNet on both complete scans and partial synthetic scans. We see that the segmentation error $E_{\text{seg}}$ of BaseNet is comparable to our method, but the axis direction error $E_{\text{angle}}$, axis position error $E_{\text{dist}}$, and motion type error $E_{\text{type}}$ of BaseNet are 48.9\% higher than ours or more. 

The main reason for this discrepancy in the results could be that segmentation and classification are an easier task than mobility prediction. Network architectures like PointNet++ have already shown good results on those two tasks, while for mobility prediction, the single input frame may lead to ambiguities in the inference. 
In our deep learning framework, RPM-Net uses an RNN to generate sequences consisting of multiple frames that describe the motion, which constrains more the inference. As a consequence, the prediction of the parameters with Mobility-Net is much more accurate.

\begin{figure}[!t]
    \centering
    \includegraphics[width=0.9\linewidth]{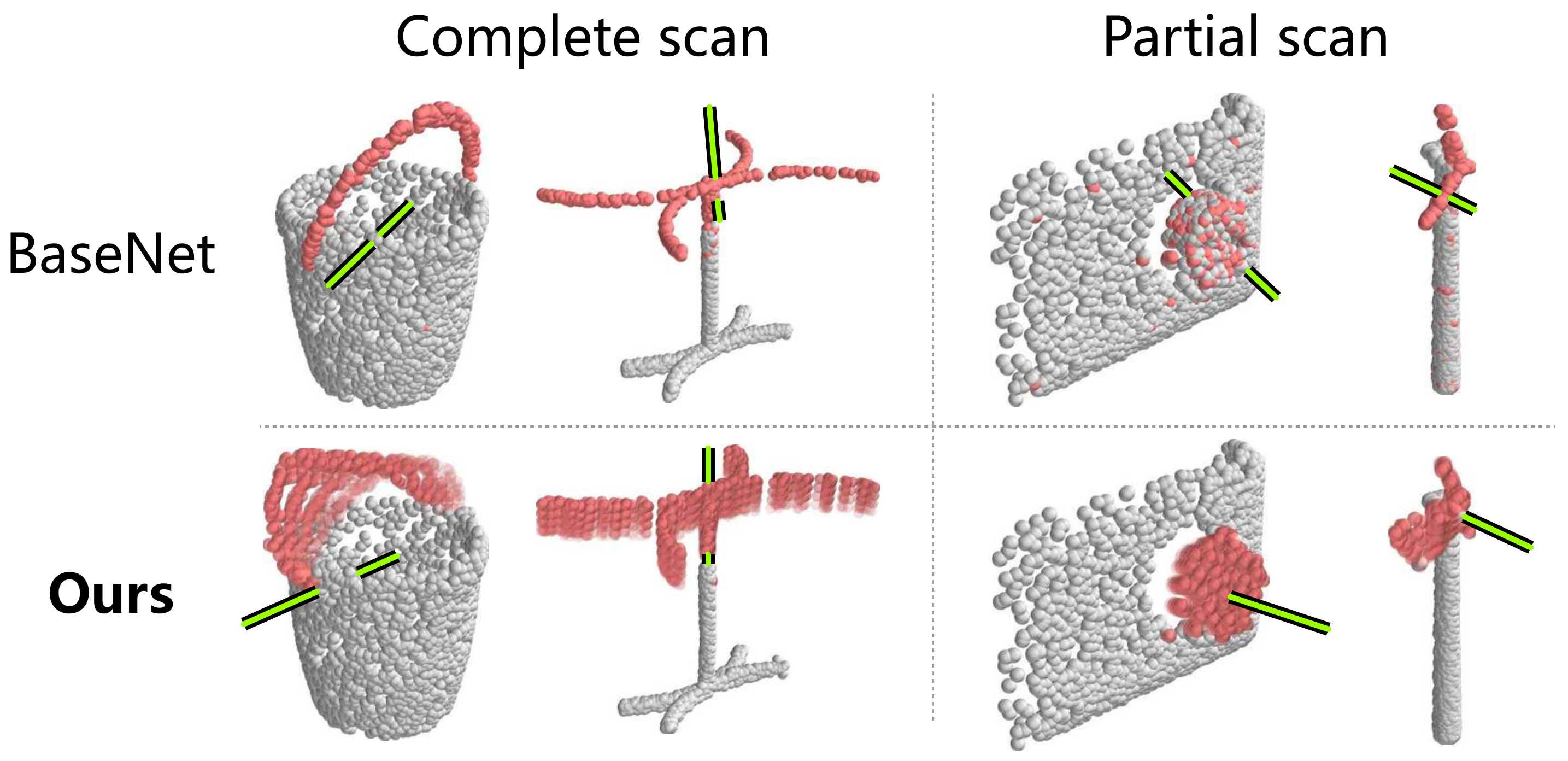}
\caption{Visual comparison of our method to BaseNet.}
\label{fig:comp_basenet}
\end{figure}

Figure~\ref{fig:comp_basenet} shows a visual comparison of our method to BaseNet on a few examples. Since BaseNet does not generate motion frames, we show its segmentation and predicted axis on the input point cloud, while for our method, we show four consecutive frames all together with the predicted segmentation and axis. The moving parts of the generated frames are shown in lighter color when they are closer to the input frame. 
We can see that BaseNet tends to provide inaccurate predictions, for example, predicting a rotation axis along the wrong direction for the handle of the bucket.

\subsubsection{Comparison to previous works.}
We compare our method to two previous works: SGPN~\cite{wang2018} and the concurrent work Shape2Motion~\cite{wang2019shape2motion} \rh{on our dataset}. Since SGPN only performs segmentation, we evaluate it with the baseline designed by Wang et al.~\shortcite{wang2019shape2motion} that takes the segmentation and uses it to estimate mobility parameters (denoted SGPN+BL). Shape2Motion directly predicts the mobility of static shapes. We perform the comparison on our dataset of complete point clouds. By inspecting the results in Table~\ref{tab:comp_shape2motion} \rh{on our dataset}, we see that we obtain the lowest errors in all the error measures, noticeably about half the error of Shape2Motion.
\begin{table}[!t]%
	\caption{
	Errors in motion-based segmentation and mobility prediction for our method and previous works (Shape2Motion (S2M)~\cite{wang2019shape2motion} and SGPN+BL~\cite{wang2018}) evaluated on two different datasets. }
	\label{tab:comp_shape2motion}
	\begin{minipage}{\columnwidth}
		\begin{center}
		\begin{tabular}{l||l||l|l|l|l}
			\hline
			\textbf{Dataset} & \textbf{Method} & $E_{\text{seg}}$ & $E_{\text{angle}}$ & $E_{\text{dist}}$ &  $E_{\text{type}}$ \\ \hline
		 \multirow{3}{*}{Ours} & S2M & 0.463  & 0.261  & 0.279 & 0.064 \\ \cline{2-6} 
		 & SGPN+BL    & 0.688   & 0.319 & 0.294 &  0.085 \\ \cline{2-6} 
		 & \textbf{Ours}& {\bf 0.205} &{\bf 0.147} &  {\bf 0.176}  & {\bf 0.019}  \\ \hline
		 \rh{\multirow{2}{*}{S2M} } & S2M & 0.272 & 0.175 & 0.192 & 0.033 \\ \cline{2-6} 
		 &\textbf{Ours}& {\bf 0.211} &{\bf 0.138} & {\bf 0.145} & {\bf 0.016}  \\ \hline
		\end{tabular}
		\end{center}
	\end{minipage}
\end{table}%

\begin{figure}[!t]
    \centering
    \includegraphics[width=0.99\linewidth]{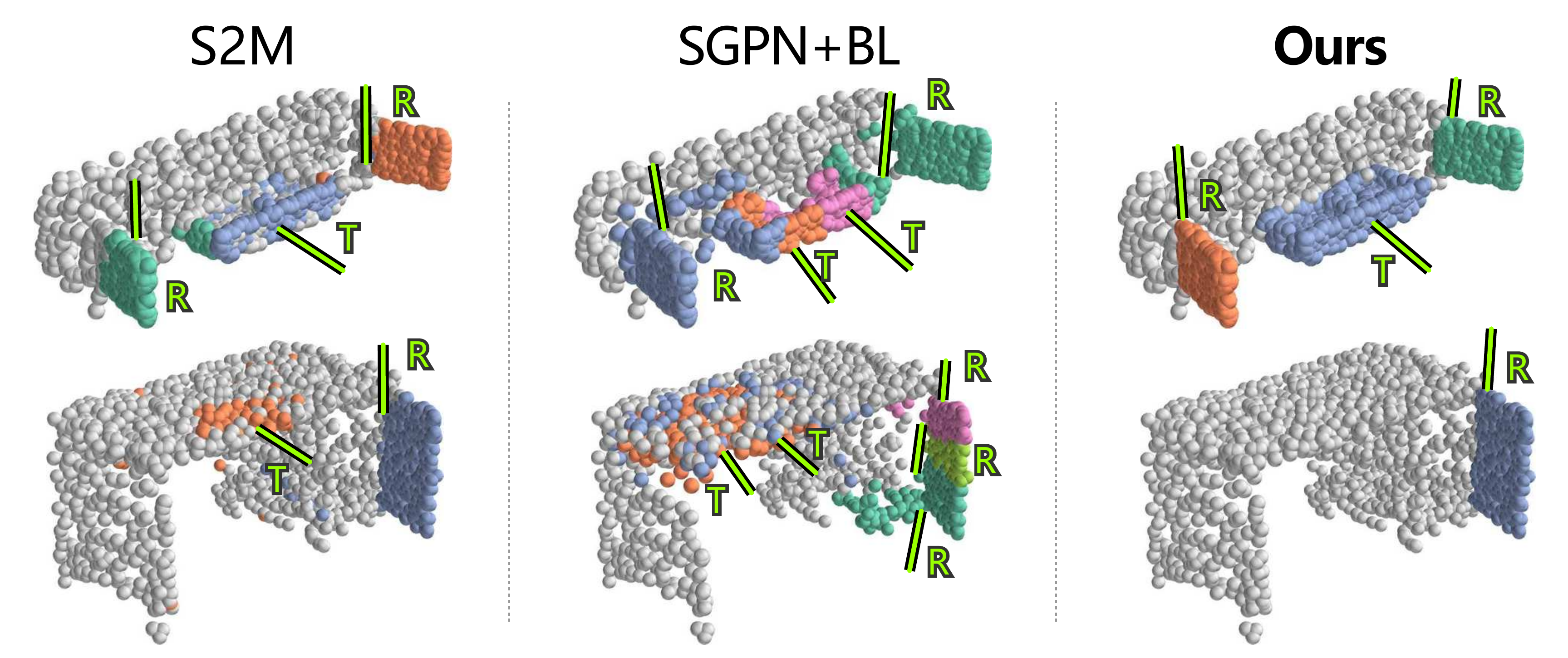}
\caption{Visual comparison of our method to previous works (Shape2Motion (S2M)~\cite{wang2019shape2motion} and SGPN+BL~\cite{wang2018}) \rh{on example shapes from our dataset}.}
\label{fig:comp_previous}
\end{figure}

Figure~\ref{fig:comp_previous} shows visual comparisons of our method to these previous works. Note how the previous methods produce noisier segmentations and detect spurious moving parts. In both examples, SGPN+BL detects multiple moving parts that should be in fact a single part. Shape2Motion displays a better performance, obtaining a cleaner segmentation, but still detects spurious parts such as the drawer on the desk of the second row, although no drawer is present in this shape. Our method provides the best results by predicting the segmentation together with the displacement maps, and then estimating the mobility parameters from the displacement maps.

\begin{figure}[!t]
    \centering
    \includegraphics[width=0.95\linewidth]{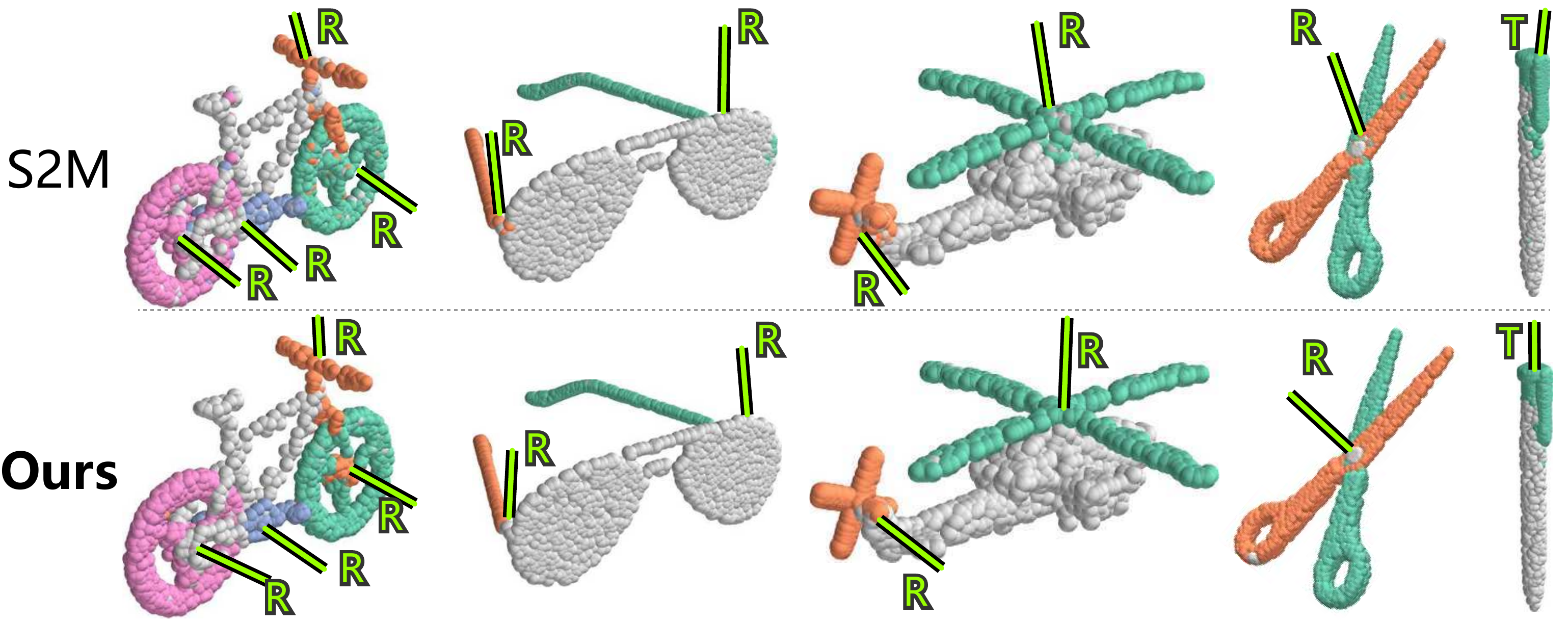}
\caption{\ok{Visual comparison of our method to Shape2Motion (S2M)~\cite{wang2019shape2motion} on example shapes from their dataset. } }
\label{fig:comp_s2m}
\end{figure}

\ok{We further compare our method to Shape2Motion~\cite{wang2019shape2motion}  on subsets of their dataset consisting of 1,885 objects belonging to 33 different categories, which is double the size of our dataset. To be able to train our networks, we added the motion range for each moving part in their dataset, to set the start and end states of the motion. 
We removed the categories of their dataset where shapes are not aligned, such as {\em swiss army knives}, categories where the entire shape moves without the existence of a static reference part, such as {\em rocking chairs}, and categories where the shapes have more than one moving part connected in a sequence, such as {\em folding table lamps}, whose start and end states are ambiguous.
We see in Table~\ref{tab:comp_shape2motion} that our method can achieve comparable performance \rh{on} Shape2Motion \rh{dataset} as on our own dataset, which shows the generality and scalability of our method. While Shape2Motion obtains better results on their own dataset, the errors are still generally larger than ours. Note that the errors of  Shape2Motion are slightly larger than what is reported in their paper. The main reason is that we generate the whole motion sequence for each object based on the specified motion parameters, and test objects in different states. In their evaluation, only one state is tested, which is likely a median case that is less ambiguous. Moreover, Figure~\ref{fig:comp_s2m} shows a visual comparison of our method to Shape2Motion on a few examples. Our method obtains better part segmentations for the bicycle and helicopter and more accurate predictions for the rotational axes of the glass and scissors \rh{and translational axis of the pen} . Note that we also have the scissors category in our dataset but with a different segmentation, as shown in the fourth row of Figure~\ref{fig:gallery}. Our method can learn to predict the right motion and corresponding segmentation in both cases. }

\subsubsection{Generality of the method on shapes with non-trivial motion}
\begin{figure}[!t]
    \centering
    \includegraphics[width=0.94\linewidth]{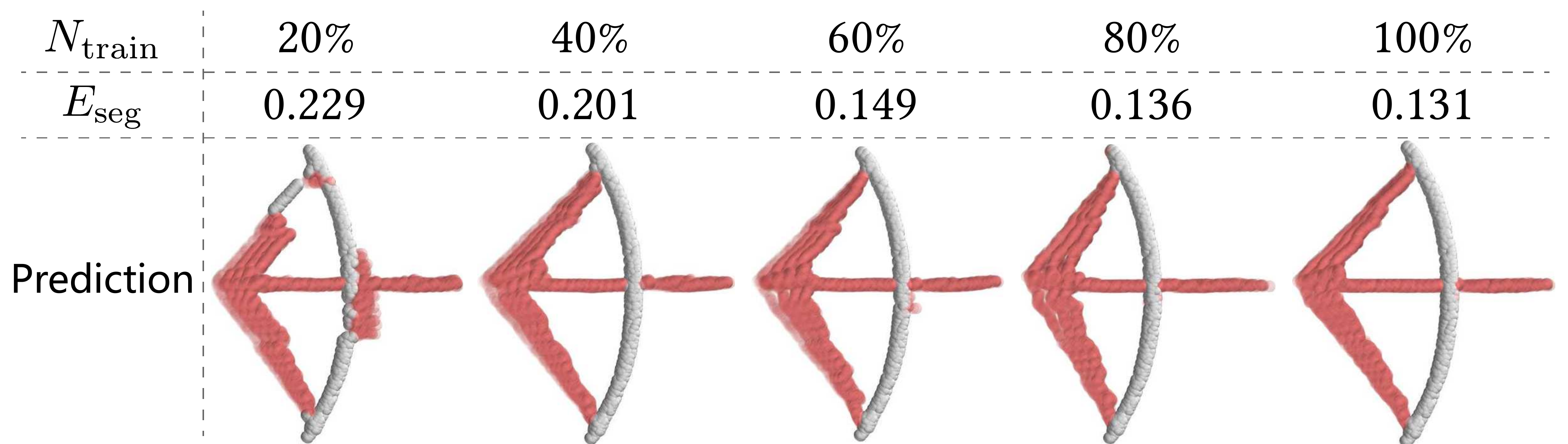}
\caption{Prediction results on shapes with non-trivial motion, when learning from increasing training sets. The first row shows the percentages of the dataset that we used for training RPM-Net. The second row shows the corresponding segmentation errors on the test sets. The last row shows visual results obtained on a selected test shape.}
\label{fig:nontrivial_testimg}
\end{figure}

\ok{For shapes with non-trivial motions, since there are no mobility parameters for which we can evaluate the prediction accuracy, we only compute $E_{\text{seg}}$ for evaluating the results. To evaluate how large the dataset should be to so that the method generalizes well, we train RPM-Net on datasets with increasing sizes and check how the prediction results change for the shapes with non-trivial motion. As we see in Figure~\ref{fig:nontrivial_testimg}, $E_{\text{seg}}$ decreases when less than 60\% of the training data is used, but becomes stable when more data is used. The same trend can also be seen in the visual examples of the prediction results, where the results obtained with 20\% and 40\% training data are not so desirable, but the results obtained with 60\% of the data or more are satisfactory, following the desired motion more closely.}

\subsection{Ablation studies}
\label{sec:ablation}

To further justify our network design, we perform five ablation studies for the motion hallucination network and two ablation studies for the mobility prediction network. To compare different versions of the method in these studies, we report the average of the error measures computed on the test set of complete scans.

\begin{table}[!t]%
	\caption{Ablation studies where we compare our RPM-Net to
        versions of the network where we remove the RNN module or selected terms of the loss function. Note the importance of all the components and loss terms in providing the lowest errors for all the error measures (last row).}
	\label{tab:comp_disp}
	\begin{minipage}{\columnwidth}
		\begin{center}
		\begin{tabular}{l||l|l|l|l}
			\hline
			 \textbf{Method} &  $E_{\text{seg}}$ & $E_{\text{angle}}$&  $E_{\text{dist}}$  &  $E_{\text{type}}$ \\ \hline
			 w/o RNN & 0.452 & 0.256 & 0.390  & 0.087  \\ \hline
		     w/o $L_{\text{geom}}$ & 0.236 & 0.151 & 0.187  & 0.028  \\ \hline
		     w/o $L_{\text{disp}}$ & 0.324 & 0.298 & 0.401  & 0.059  \\ \hline
		     w/o $L_{\text{mot}}$  & 0.254 & 0.199 & 0.280  & 0.031  \\ \hline
 		     w/o $L_{\text{seg}}$  & 0.785 & 0.772 & 0.625  & 0.124  \\ \hline
		     \textbf{Ours}   & {\bf 0.205}  &{\bf 0.147} &  {\bf 0.176} & {\bf 0.019}  \\ \hline
		\end{tabular}
		\end{center}
	\end{minipage}
\end{table}%

\subsubsection{Importance of RNN}
Using an RNN is important for generating dynamic sequences that illustrate the motion of the input objects. To justify the benefit of generating a motion sequence, we compare our RPM-Net to a version of the network that only generates one displacement map with the LSTM and then combines it with the input cloud to infer the segmentation and mobility parameters. The comparison can be seen by contrasting the first and last rows in Table~\ref{tab:comp_disp}. We see that a motion sequence hallucinated by our method provides a more accurate segmentation and parameter prediction than when only a single frame is generated.

\subsubsection{Importance of $L_{\text{rec}}$}
To show the importance of $L_{\text{rec}}$ consisting of $L_{\text{geom}}$ and $L_{\text{disp}}$, which are the terms of the loss function comparing the predicted point clouds $\mathcal{P}_t$ and displacement maps $\mathcal{D}_t$ to the ground-truth, we compare the result of our method to results obtained without adding either of those two terms. We report the error values obtained in this experiment in the second and third rows of Table~\ref{tab:comp_disp}, compared to using our full loss function in the last row. We see that $L_{\text{disp}}$ is quite important in lowering the error rates for all the measures. Although the error values of mobility prediction without using $L_{\text{geom}}$ are close to the values when using the full loss function, the importance of this term is demonstrated when measuring the quality of the segmentation $E_{\text{seg}}$.

\begin{figure}[!t]
    \centering
    \includegraphics[width=0.99\linewidth]{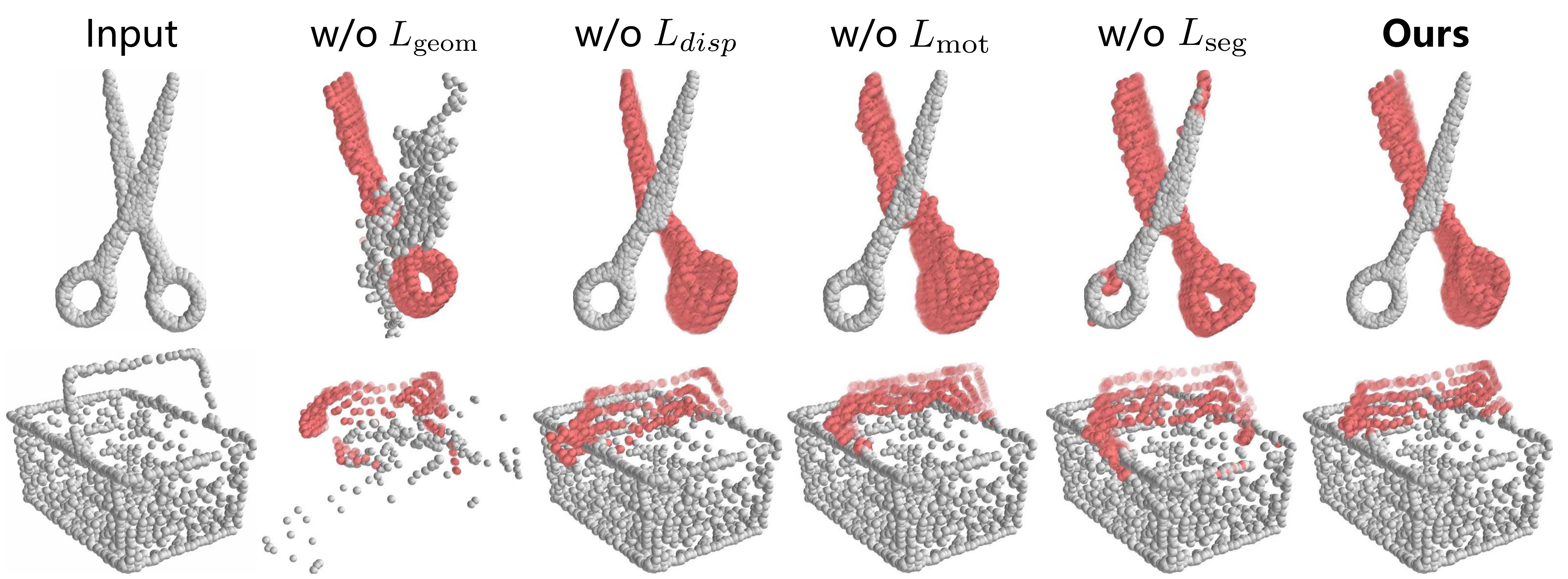}
\caption{Ablation experiments of our method. The reference part is colored gray while multiple frames of the moving part are colored in red.}
\label{fig:ablation}
\end{figure}

In addition, when inspecting the visual example shown in Figure~\ref{fig:ablation}, we can see that without $L_{\text{geom}}$, the points, especially those in the reference part, tend to move into unexpected locations, although the motion of the moving part looks reasonable thanks to the effect of the displacement loss $L_{\text{disp}}$. On the other hand, when $L_{\text{disp}}$ is removed, the motion of the points on the moving part becomes inconsistent, which results in  distortion of the moving part. In comparison, our full method can predict a correct and smooth motion for the moving part and also keep the reference part unchanged.

\subsubsection{Importance of $L_{\text{mot}}$} 
Table~\ref{tab:comp_disp} and Figure~\ref{fig:ablation} also show the importance of the motion loss in obtaining high-quality motion hallucination results. When the term is removed, the average errors increase, and the visual example in Figure~\ref{fig:ablation} shows that the lack of this term in the loss leads to motions that are less smooth, reflected by the rough appearance of the predicted frames.

\subsubsection{Importance of $L_{\text{seg}}$} 
To show the importance of the $L_{\text{seg}}$ term in the loss, we train a version of the network without this term. To obtain the segmentation in this case, we filter points depending on whether they move more than an appropriate threshold $\theta$ in the displacement maps $\mathcal{D}_t$, to segment the points into moving and static (reference) points. In our experiments, we use a threshold $\theta = 0.01$ for determining the segmentation. We see in Table~\ref{tab:comp_disp} that without $L_{\text{seg}}$ the average errors increase significantly. When inspecting the visual example in Figure~\ref{fig:ablation}, we see that the lack of the segmentation loss leads to clear errors in the segmentation, especially for points at the extremity of the scissors in this example.

\begin{table}[t!]%
	\caption{
		Ablation studies where we compare Mobility-Net to a method which estimates the parameters directly from the displacement fields instead of using neural networks (denoted ``w/o $L_{\text{mob}}$''), and to a version of the network where we remove the point cloud $\mathcal{P}_0$ from the input (denoted ``w/o $\mathcal{P}_0$''). Note that our method provides the lowest errors (last row).}
	\label{tab:comp_input}
	\begin{minipage}{\columnwidth}
		\begin{center}
		\begin{tabular}{l||l|l|l}
			\hline
			 \textbf{Method} & $E_{\text{angle}}$&  $E_{\text{dist}}$ &   $E_{\text{type}}$ \\ \hline
		  w/o $L_{\text{mob}}$ & 0.722 &  0.684 &  0.115 \\ \hline
		  w/o $\mathcal{P}_0$  & 0.569 &  0.631 &   0.097 \\ \hline
		   \textbf{Ours}   &{\bf 0.147} &  {\bf 0.176} & {\bf 0.019}  \\ \hline
		\end{tabular}
		\end{center}
	\end{minipage}
\end{table}%

\subsubsection{Importance of $L_{\text{mob}}$}
We justify the use of the mobility loss $L_{\text{mob}}$ by comparing our Mobility-Net to a method that algorithmically infers the mobility parameters $\mathcal{M}$ from the predicted displacement maps. Specifically, we generate a point cloud motion sequence $\mathcal{P}_t$ from the displacement maps $\mathcal{D}_t$. Then, we compute the optimal rigid transformation matrix with minimum mean square error that transforms one frame into another, and extract the axis direction for translations, and axis direction and position for rotations and the combination of translation and rotation. 
We report the error values for this experiment in the first row of Table~\ref{tab:comp_input}.

\begin{figure}[!t]
    \centering
     \includegraphics[width=0.8\linewidth]{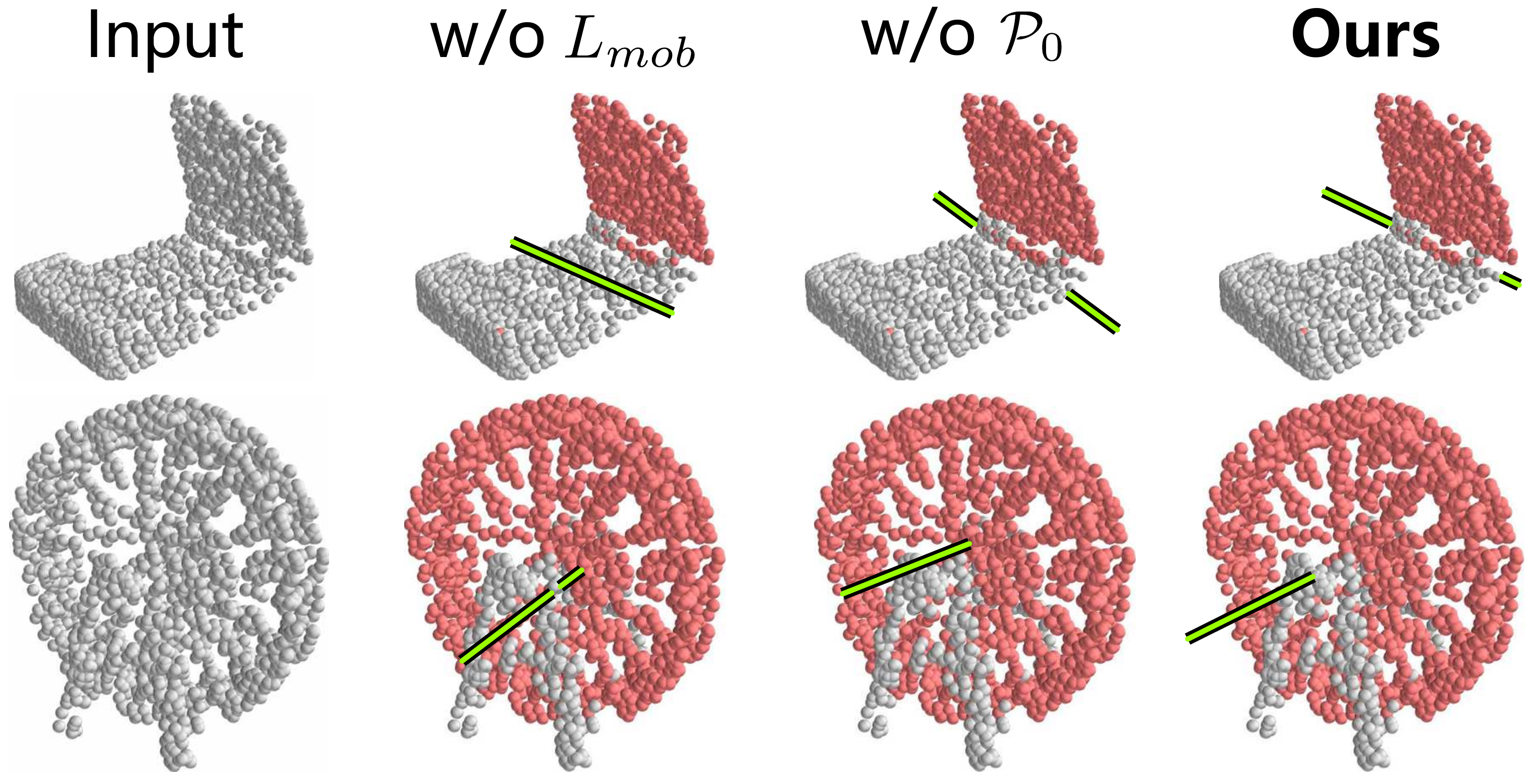}
\caption{Visual comparison of our method to results where the mobility parameters are not obtained by network prediction  (denoted ``w/o $L_{\text{mob}}$'') or the input point cloud was removed from the input (denoted ``w/o $\mathcal{P}_0$''). }
\label{fig:ablation_motion}
\end{figure}

By analyzing the errors in Table~\ref{tab:comp_input}, we see that this motion fitting approach is quite sensitive to noise, leading to large errors, while the prediction obtained with our full network is more stable and provides better results. 
Examples of the motion parameter fitting results compared to our results are shown in Figure~\ref{fig:ablation_motion}. 
We see that,  without the mobility loss $L_{\text{mob}}$, the noise in the displacements of different points can also cause large errors in the axis fitting, as seen by the incorrect location for the rotational axis of the flip phone, or the incorrect angle for the axis of the ferris wheel.

\subsubsection{Importance of $\mathcal{P}_0$}
The second row of Table~\ref{tab:comp_input} shows that, when we only provide the displacement maps and segmentation as input to Mobility-Net, the mobility prediction leads to large errors. Providing the point cloud $\mathcal{P}_0$ as additional input improves the accuracy considerably as shown in the last row of the table. 
Examples of results obtained without $\mathcal{P}_0$ compared to our results are also shown in Figure~\ref{fig:ablation_motion}. We see in the two examples that the predicted axes have noticeable errors in their direction when $\mathcal{P}_0$ is not given as input. We conjecture that, when only displacements maps are given as input to the network, these maps have small errors which accumulate and yield large errors in the regression of the transformation axis. Providing the point cloud as input helps to anchor the geometry of the shape and yield a better prediction of the axis.

\section{Conclusion and future work} \label{sec:future}

We introduced RPM-Net, a recurrent deep network that predicts the motion of shape parts, effectively partitioning an input point cloud into a reference part and one or more moving parts. We also introduced Mobility-Net, a deep network that predicts high-level mobility parameters that describe the potential motion of the moving parts detected by RPM-Net. The networks are trained on a dataset of objects annotated with ground-truth segmentations and motion specifications. However, once trained, the networks can be applied to predict the motion and mobility of a single unsegmented point cloud representing a static state of an object. We demonstrate with a series of experiments that our networks can infer the motion of a variety of objects with diverse mobility and coming from various sources, including complete objects and raw partial scans.

\begin{figure}[!t]
    \centering
    \includegraphics[width=0.95\linewidth]{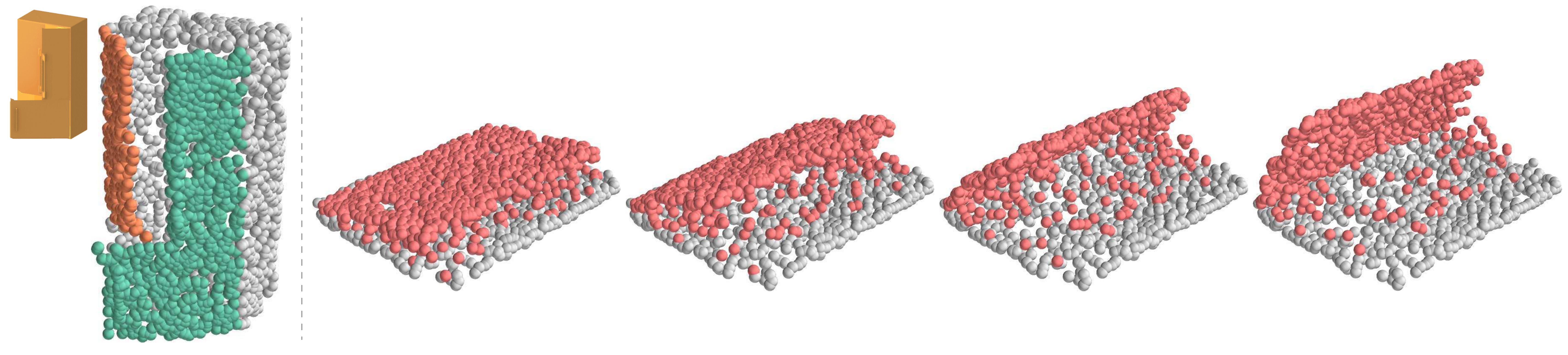}
\caption{\ok{Example failure cases of motion prediction with RPM-Net. Left: a cabinet with incorrect segmentation on the doors. Right: a closed laptop with imperfect generated motion.}}
\label{fig:failure}
\end{figure}

As demonstrated in the experiments, our method displays a high accuracy for predicting the motion of objects with one or more moving parts, where all the parts are connected to a single reference part. In addition, we showed the potential of our method in detecting hierarchical motions where multiple moving parts are connected in a sequence (Figure~\ref{fig:multi_level}), and non-trivial motions that cannot be described with a small set of transformation parameters (Figure~\ref{fig:nontrivial_motion}). 
However, \ok{as shown by Figure~\ref{fig:failure}, RPM-Net can also provide incorrect predictions when the input geometry is ambiguous. For example, the network may produce incorrect part segmentations when two parts with the same type of motion are well-aligned, like the \rh{two} doors of a cabinet \rh{that open with the same range shown in green}, or generate imperfect motion sequences with outlier points when moving and reference parts are spatially too close.}

\ok{Another limitation is that our method currently is not able to hallucinate forward and reverse motions together, since adding the training sequences in both directions will cause ambiguity. One possibility to address this limitation is to train another network for predicting reverse motions separately, and combine the prediction from the forward and reverse networks for predicting the full range of motion for shapes given in any state. It would also be interesting to explore more sophisticated solutions to this problem.}

Further experiments are needed to quantitatively evaluate our method for these more complex tasks, which would also require curating a dataset of objects with hierarchical motions and non-trivial motions, and their prescribed segmentations and ground-truth displacement maps. In addition, our current dataset is relatively small, composed of 969 object, and it assumes that the shapes are meaningfully oriented. Another more immediate direction to improve the applicability of the method to more complex scenarios is to augment our dataset by applying random transformations to the shapes, so that our network can operate in a pose-invariant manner, or to train our network with partial scans to further improve its robustness.

Another direction for future work is to leverage the mobility predicted by our method to synthesize motions for the input shapes. As part of this motion synthesis problem, one interesting subproblem is to learn how to complete the geometry of objects which may be missing when a motion takes place, e.g., a drawer being pulled from a cabinet should reveal its interior, which will be missing if the shape was scanned or did not have its interior modeled. A possible approach would be to learn how to synthesize the missing geometry from the predicted motion and existing part geometry. Such an approach would require at the minimum a training set in the form of segmented objects with their interiors captured.

\begin{acks}
We thank the reviewers for their valuable comments. This work was supported in parts by 973 Program (2015CB352501), NSFC (61872250, 61602311, 61861130365),  GD Science and Technology Program (2015A030312015), LHTD (20170003), Shenzhen Innovation Program (JCYJ20170302153208613), NSERC (611370, 2015-05407), Adobe gift funds, and the National Engineering Laboratory for Big Data System Computing Technology.
\end{acks}

\bibliographystyle{ACM-Reference-Format}
\bibliography{mobility}

\end{document}